\documentclass[twoside,11pt]{article}

\usepackage{jmlr2e}
\usepackage{comment}
\usepackage{amsmath,amssymb}
\usepackage{bm}
\usepackage{algorithm}
\usepackage[noend]{algpseudocode}
\usepackage{xcolor}
\usepackage[section]{placeins}

\newcommand{\SNR}{{\text{SNR}}}
\newcommand{\Var}{{\text{Var}}}
\newcommand{\mtry}{\texttt{mtry}}
\newcommand{\maxnodes}{\texttt{maxnodes}}

\newcommand{\norm}[1]{\left\lVert#1\right\rVert}
\newcommand{\bt}[1]{\hat{\bm{\beta}}^{#1}}

\DeclareMathOperator{\argmin}{argmin}

\usepackage{lastpage}
\jmlrheading{21}{2020}{1-\pageref{LastPage}}{10/19; Revised
7/20}{8/20}{19-905}
\ShortHeadings{Randomization as Regularization}{Mentch and Zhou}

\begin{document}

\title{Randomization as Regularization:  A Degrees of Freedom Explanation for Random Forest Success}

\author{\name Lucas Mentch \email lkm31@pitt.edu \\
	\name Siyu Zhou \email siz25@pitt.edu \\
       \addr Department of Statistics\\
       University of Pittsburgh\\
       Pittsburgh, PA 15260, USA}

\editor{Isabelle Guyon}

\maketitle

\begin{abstract}
Random forests remain among the most popular off-the-shelf supervised machine learning tools with a well-established track record of predictive accuracy in both regression and classification settings.  Despite their empirical success as well as a bevy of recent work investigating their statistical properties, a full and satisfying explanation for their success has yet to be put forth. Here we aim to take a step forward in this direction by demonstrating that the additional randomness injected into individual trees serves as a form of implicit regularization, making random forests an ideal model in low signal-to-noise ratio (SNR) settings. Specifically, from a model-complexity perspective, we show that the mtry parameter in random forests serves much the same purpose as the shrinkage penalty in explicitly regularized regression procedures like lasso and ridge regression.  To highlight this point, we design a randomized linear-model-based forward selection procedure intended as an analogue to tree-based random forests and demonstrate its surprisingly strong empirical performance.  Numerous demonstrations on both real and synthetic data are provided.  \\
\end{abstract}


\section{Introduction}
\label{sec:intro}
Despite being proposed nearly two decades ago, random forests \citep{Breiman2001} remain among the most popular and successful off-the-shelf supervised learning methods with demonstrated success in numerous scientific domains from ecology \citep{Prasad2006,Cutler2007,Coleman2017} to image recognition \citep{Bernard2007,Huang2010,Guo2011,Fanelli2013} to bioinformatics \citep{Diaz2006,Mehrmohamadi2016}.  In comparison with many competing methods, random forests are generally quite competitive in terms of predictive accuracy despite being largely considered simpler and more computationally tractable.  A recent large-scale empirical study by \cite{Fernandez2014} found them to be the top performing classifier when compared across 100's of alternatives across 100's of datasets.  Though Breiman once famously crowned Adaboost \citep{Freund1996} the best ``off-the-shelf classifier in the world", random forests have since taken that spot in the minds of many. 

The sustained success of random forests has led naturally to the desire to better understand the statistical and mathematical properties of the procedure.  \cite{Lin2006} introduced the potential nearest neighbor framework and \cite{Biau2010} later established related consistency properties.  \cite{Biau2008}, \cite{Biau2012}, and \cite{Klusowski2019sharp} analyzed convergence properties and rates for idealized versions of the procedure.  Methods for obtaining estimates of the standard errors of random forest predictions were developed in \cite{Sexton2009} and \cite{Wager2014} while algorithmic convergence studies determining how many trees are necessary for stabilization were provided by \cite{Lopes2019regression} and \cite{Lopes2019classification}.  The out-of-bag variable importance measures originally suggested by \cite{Breiman2001} have since been analyzed thoroughly and shown to be biased towards correlated features and those with many categories \citep{Strobl2007,Strobl2008,Tolocsi2011,Nicodemus2010}.  \cite{Hooker2019} recently provided an explanation for this kind of behavior.  The core random forest methodology has been extended to numerous frameworks including quantile regression \citep{Meinshausen2006}, reinforcement learning \citep{Zhu2015}, and survival analysis \citep{Hothorn2005,Ishwaran2008,Cui2017,Steingrimsson2019}.  For a more comprehensive overview of general random-forest-related research, we refer the reader to a recent review paper from \cite{Biau2016}. 

In the last several years, a number of important statistical properties of random forests have also been established whenever base learners are constructed with subsamples rather than bootstrap samples.  \cite{Scornet2015} provided the first consistency result for Breiman's original random forest algorithm whenever the true underlying regression function is assumed to be additive.  \cite{Mentch2016} provided the first distributional results for random forests, demonstrating that predictions are asymptotically normal under various regularity conditions and proposing accompanying procedures for producing confidence intervals and hypothesis tests for feature importance.  \cite{Mentch2017} extended this idea to grid-structured testing to allow for hypothesis tests for additivity in the regression function.  \cite{Coleman2019} proposed a more computationally efficient permutation-based hypothesis test allowing such procedures to scale easily to big data settings where large test sets may be utilized.  \cite{Wager2018} established both consistency and asymptotic normality for \emph{honest} and \emph{regular} trees whenever many trees are constructed.  \cite{Peng2019} weakened the conditions necessary for asymptotic normality in a more general setting and provided Berry-Esseen Theorems describing their rate of convergence. 

Despite the impressive volume of research from the past two decades and the exciting recent progress in establishing their statistical properties, a satisfying explanation for the sustained empirical success of random forests has yet to be provided.  In their recent review, \cite{Biau2016} note that research investigating the properties of random forest tuning parameters is ``unfortunately rare" and that ``present results are insufficient to explain in full generality the remarkable behavior of random forests."  Indeed, while numerous studies have experimented with various tuning parameter setups and strategies \citep{Genuer2008,Bernard2009,Genuer2010,Duroux2016,Scornet2017,Probst2017,Probst2019}, results from these works have largely been limited to high-level, somewhat heuristic takeaways with most concluding only that building more trees helps to stabilize predictions and that the other tuning parameters can have a substantial effect on accuracy.  Other work has attempted to garner insight by drawing connections between tree-based ensembles and other related procedures such as kernel methods \citep{Scornet2016,Olson2018} or neural networks \citep{Welbl2014,Biau2016neural}.  However, to our knowledge, besides the original motivation provided by \cite{Breiman2001}, the only other work devoted to attempting to explain the success of random forests was provided recently by \cite{Wyner2017} who hypothesized that their strong performance was due to their behavior as ``self-averaging interpolators."  \cite{Belkin2019} very recently wrote in support of this notion.  However, as expanded upon in detail in the following section, in our view, these existing explanations fall short on a number of fronts.  In general, the arguments provided apply equally well to other learning methods not generally thought to have strong and robust empirical performance, fail to provide insight into the role played by the randomness, and apply only to particular problem setups.   

In contrast, the work presented here offers a concrete explanation for the role played by the extra randomness most commonly injected into the base learners in random forest procedures.  Instead of assuming that random forests simply do ``work well", we take a more principled approach in trying to isolate the effects of that additional randomness and determine when its inclusion results in improved accuracy over a baseline approach like bagging that uses non-randomized base learners.  In particular, we argue that the additional randomness serves to regularize the procedure, making it highly advantageous in low signal-to-noise ratio settings.  Speculation along these lines was hypothesized informally in \cite{esl} who observe that random forests sometimes behave similarly to ridge regression.  


To drive home this point, we further demonstrate that incorporating similar randomness into alternative (non tree-based) model selection procedures can result in improved predictive accuracy over existing methods in exactly the settings where such improvements would be expected.  In particular, inspired by recent work on degrees of freedom for model selection by \cite{Tibshirani2015} and \cite{Hastie2017}, we consider two randomized forward selection procedures for linear models designed as analogues to classical bagging and random forests and demonstrate the same kind of regularization properties. Our findings in this setting are thus similar in spirit to those produced by \cite{Wager2013} who demonstrate a regularization effect arising from dropout training applied to generalized linear models. 

The remainder of this paper is laid out as follows.  In Section \ref{sec:RFexplanations} we formalize the random forest procedure and continue the above discussion, providing something of a literature review of recent random forest analyses as well as a more detailed overview of the shortcomings of existing explanations for their success.  In Section \ref{sec:RFdof} we discuss degrees of freedom for model selection procedures and demonstrate that within a traditional random forest context, more randomness results in procedures with fewer degrees of freedom.  We emphasize and build upon this finding in Section \ref{sec:RFsnr} by demonstrating in numerous settings using both real and synthetic data that the relative improvement in accuracy offered by random forests appears directly related to the relative amount of signal contained within the data.  Finally, in Section \ref{sec:RFS} we introduce the linear model forward-selection-style analogues for bagging and random forests and produce near identical results, finding in particular that in noisy low-dimensional settings, injecting randomness into the selection procedure can outperform even highly competitive explicit regularization methods such as the lasso.  Example code for the simulations and experiments presented is available at \url{https://github.com/syzhou5/randomness-as-regularization}.

\section{Random Forests and Existing Explanations}
\label{sec:RFexplanations}

We begin by formalizing the random forest framework in which we will work in the following sections.  Unless otherwise noted, throughout the remainder of this paper we will consider a general regression framework in which we observe (training) data of the form $\mathcal{D}_n = \{Z_1, ..., Z_n\}$ where each $Z_i = (\bm{X}_i,Y_i)$, $\bm{X}_i = (X_{1,i}, ..., X_{p,i})$ denotes a vector of $p$ features, $Y \in \mathbb{R}$ denotes the response, and the variables have a general relationship of the form
\begin{equation}
Y = f(X)+\epsilon
\label{eqn:reg}
\end{equation}

\noindent where $\epsilon$ is often assumed to be independent noise with mean 0 and variance $\sigma^{2}$.  For a given point $z = (\bm{x},y)$, a random forest prediction at $\bm{x}$ takes the form
\begin{equation}
\hat{y} = \text{RF}(\bm{x}; \mathcal{D}_n, \Theta) = \frac{1}{B} \sum_{i=1}^{B} T(\bm{x}; \mathcal{D}_n, \Theta_i)
\label{eqn:rf}
\end{equation}

\noindent where the base-learners $T$ are typically tree-based models constructed on some resample of the original data $\mathcal{D}_n$ and the randomness involved in the procedure is indexed by $\Theta_i$.  In the work that follows we assume these base models are regression trees built according to the standard CART criterion \citep{CART} whereby an internal node $t$ is split in an axis-aligned fashion into left and right daughter nodes of the form $t_L = \{\bm{x} \in t: \, x_j \leq s\}$ and $t_R = \{\bm{x} \in t: \, x_j > s\}$ whenever the decision is made to split the feature $X_j$ at $s$.  The particular variable and split location are chosen from among those available as that pair which minimizes the resulting within-node variance of the offspring.  For a more detailed discussion of this procedure, we refer the reader to recent work by \cite{Scornet2015} or \cite{Klusowski2019}.

Throughout the literature on random forests, it is common to succinctly contain all randomness in the single term $\Theta_i$ as in equation (\ref{eqn:rf}) above.  We note however that for our purposes below, it may be convenient to consider this more explicitly as $\Theta_i = (\Theta_{\mathcal{D},i},\Theta_{mtry,i})$.  Written in this form, $\Theta_{\mathcal{D},i}$ serves to select the resample of the original data utilized in the $i^{th}$ tree.  While much recent work has focused on subsampled random forests, here we consider the $B$ resamples to be bootstrap samples as originally put forth in \cite{Breiman2001}.  The second randomization component $\Theta_{mtry,i}$ then determines the eligible features to be split at each node in the $i^{th}$ tree.  As is standard, we assume that at each internal node, the split must be chosen from among only $\texttt{mtry} \leq p$ candidate features and that such candidates are selected uniformly at random without replacement.  When $\mtry = p$, the procedure reduces to bagging \citep{Breiman1996}.

\subsection{Explanations for Random Forest Success}
The original reasoning behind random forests provided by \cite{Breiman2001} was based on an extension of the randomized tree analysis given in \cite{Amit1997}.  Breiman showed that the accuracy of any randomized ensemble depends on two components:  the strength (accuracy) of the individual base-learners and the amount of dependence between them.  Thus, the original motivation for a procedure like random forests might be seen from a statistical perspective as akin to the classic bias-variance tradeoff.  In the same way that some procedures (e.g.\ the lasso \citep{Tibshirani1996,Chen2001}) consider trading a small amount of bias in exchange for a large reduction in variance, random forest ensembles might be seen as trading a small amount of accuracy at the base-learner level (by injecting the extra randomness) for a large reduction in between-tree correlation. \cite{esl} provide a thorough, high-level discussion of this effect in showing that the \texttt{mtry} parameter serves to reduce the variance of the ensemble.

However, this discussion from \cite{Breiman2001} might be better seen as motivation for why a randomized ensemble could \emph{potentially} improve accuracy rather than an explanation as to why random forests in particular \emph{do} seem to work well.  Breiman himself experiments with different kinds of randomness in the original manuscript and suggests that in practice users can also experiment with different forms to try and determine what works best in particular settings.  Furthermore, in his concluding remarks, Breiman notes that while the additional randomness at the base learner level helps to reduce the variance of the ensemble, the magnitude of improvement often seen with random forests suggested to him that perhaps it somehow also ``act[s] to reduce bias" but that ultimately ``the mechanism for this [was] not obvious."  In the years since, it has been shown quite clearly that the benefits sometimes seen with random forests are the result of variance reduction alone; see \cite{esl} for a more complete discussion.

In recent work, \cite{Wyner2017} take a more definitive stance, conjecturing that both random forests and AdaBoost \citep{Freund1996} work well because both procedures are ``self-averaging interpolators" that fit the training data perfectly while retaining some degree of smoothness due to the averaging.  The key to their success, they argue, is that in practice, datasets often contain only small amounts of noise and these algorithms are able to mitigate the effects of noisy data points by localizing their effect so as to not disturb the larger regions where the data consists mostly of signal.  Indeed, the authors acknowledge that the procedures ``do in fact overfit the noise -- but only the noise.  They do not allow the overfit to metastasize to modestly larger neighborhoods around the errors."

\subsection{Random Forests and Interpolation}
\label{subsec:rfInt}
As the central claim of \cite{Wyner2017} is that random forests ``work not in spite, but because of interpolation" we now make this notion and argument explicit.

\begin{definition}[Interpolation]\label{def:interpolating}
A classifier (or regressor) $\hat{f}$ is said to be an \emph{interpolating} classifier (regressor) if for every training point $(\bm{x}_j,y_j) \in \mathcal{D}_n$, $\hat{f}(\bm{x}_j) = y_j$.  
\end{definition}

\noindent This definition of an interpolating classifier is taken directly from \cite{Wyner2017}; for completeness and because it will be directly relevant to the immediate conversation, we expand the definition so as to apply in the same fashion to regression contexts.  

Consider first the classification setting wherein the response $Y \in \{a_1, ..., a_k \}$ and we seek to utilize the training data $\mathcal{D}_n$ to construct a classifier $\hat{f}_n: \mathcal{X} \mapsto \{a_1, ..., a_k \}$.  The random forest procedure laid out above for regression can be easily amended to perform classification -- indeed, the original procedure provided in \cite{Breiman2001} was discussed largely for classification problems.  Here, split locations within trees are chosen as those which provide the greatest empirical reduction in Gini impurity and final predictions are taken as the \emph{majority vote} -- the class predicted most often -- across all trees in the forest.  

\begin{figure}
	\centering
	\includegraphics[width=0.98\textwidth]{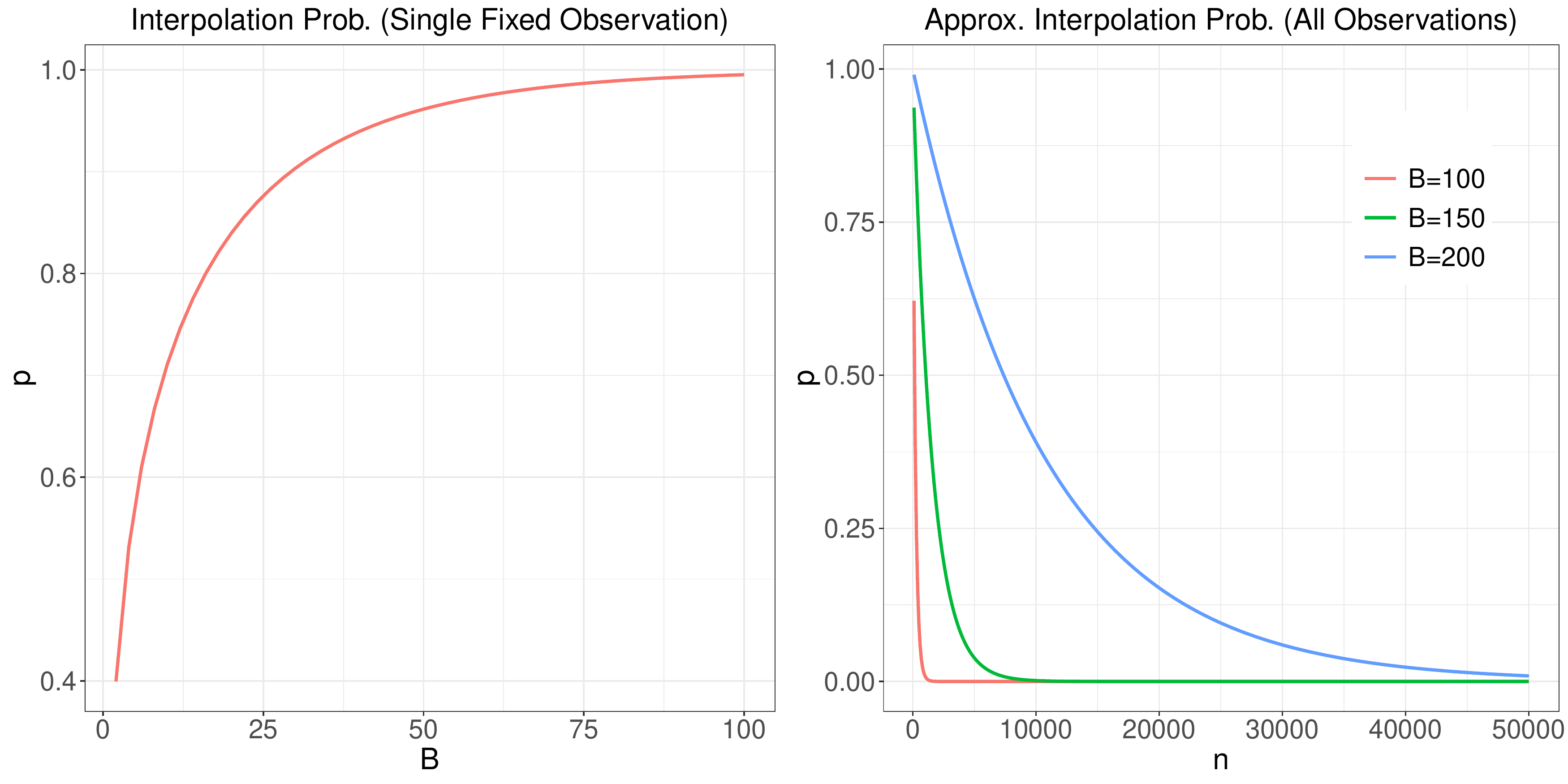}
	\caption{(Left):  Interpolation probability vs.\ number of trees ($B$) for a single observation.  (Right): Approximate interpolation probabilities vs.\ sample sizes for random forest classifiers built with different numbers of trees. Both plots pertain to the binary classification setting.}
	\label{fig:IntProb}
\end{figure}

Now consider a particular observation $z=(\bm{x},y) \in \mathcal{D}_n$.  In order to be more explicit and without loss of generality, suppose that $y=a_1$.  Given $B$ resamples of the original data $\mathcal{D}_{1}^{*}, ..., \mathcal{D}_{B}^{*}$, whenever trees are fully grown so that each terminal node contains only a single observation, it must necessarily be the case that $T(\bm{x}; \mathcal{D}_{i}^{*},\Theta_i) = a_1$ (i.e.\ the tree predicts the correct class $a_1$ for $\bm{x}$) whenever $(\bm{x},y) \in \mathcal{D}_{i}^{*}$.  Thus, in order for the random forest classifier to select the correct class for $\bm{x}$, (i.e.\ $\text{RF}(\bm{x})=a_1$) it suffices to ensure that $(\bm{x},y)$ is selected in a plurality (or simple majority in the case of binary classification) of the resamples.  When bootstrap samples are used, it is well known that the probability of each observation appearing is approximately 0.632.  Thus, given $B$ bootstrap samples, the probability that the observation $(\bm{x},y)$ appears in at least half of these resamples is approximately
\[
p_{int}(B) = 1-Bin(B/2; \, n=B, p=0.632)
\]
\noindent where $Bin(z;n,p)$ denotes the binomial cdf evaluated at the point $z$ with parameters $n$ (the number of trials) and $p$ (the probability of success in each trial).  For even moderately large $B$, this probability is quite large; see the left plot in Figure \ref{fig:IntProb}.  Thus, for binary classification problems, the probability of interpolating any given training observation is large whenever $B$ is also moderately large.

Note however that according to the definition above, a classifier is only designated as an interpolator if it interpolates \emph{all} training observations.  While an exact calculation for the probability of all $n$ points appearing in at least half of the bootstrap samples is somewhat involved, we can approximate it with $(p_{int}(B))^n$, the calculation that would result if the interpolation probabilities were independent for each observation.  Plots of this quantity are shown in the right plot in Figure \ref{fig:IntProb} across a range of sample sizes for $B = 100, 150, \text{and } 200$.  In each case, for fixed $B$, the (approximate) probability that the classifier is an interpolator tends to 0 as $n \rightarrow \infty$, suggesting that in order for random forest classifiers to necessarily interpolate, the number of bootstrap replicates $B$ must be treated as an increasing function of $n$.  Though perhaps obvious, we stress that this is not generally the manner in which such estimators are constructed.  In all software with which we are familiar, default values of $B$ are set to a fixed moderate size, independent of the size of the training set.  The \texttt{bagging} function in the \texttt{ipred} package in \texttt{R} \citep{ipred}, for example, takes 25 bootstrap replicates by default. Recent work from \cite{Lopes2019classification} has also provided a means of estimating the algorithmic variance of random forest classifiers and shown that it sometimes vanishes quite quickly after relatively few bootstrap samples.  Thus, while it's possible to construct random forests in such a way that they necessarily interpolate with high probability in classification settings, it is not clear that random forests would generally be constructed in this fashion in practice and thus it is also not clear that the interpolation-based explanation offered by \cite{Wyner2017} is sufficient to explain the strong performance of random forests, even in specific contexts.  It is also worth noting that on certain datasets where random forests happen to produce good models with low generalization error, they may likely also fit quite well on the training data, perhaps even nearly interpolating.  This, however, is certainly possible for any modeling procedure and thus in no way would aid in explaining the particular success of random forests.

\subsection{Shortcomings of Current Explanations}

Before continuing with our critique, it's worth pausing to note where the existing explanations are in agreement.  Both \cite{Breiman2001} and \cite{Wyner2017} seem to largely agree on the following points:

\begin{enumerate}
\item  Random forests and boosting behave in a very similar fashion and thus their success should be able to be explained in a very similar fashion [\cite{Breiman2001} pages 6, 20, Section 7; \cite{Wyner2017} entire paper].  
\item Random forests and boosting generally outperform most other competing methods (e.g.\ \cite{Dietterich2000} and \cite{Breiman2000}) in terms of minimizing generalization error and substantially outperform bagging  [\cite{Breiman2001} page 10; \cite{Wyner2017} page 3].    
\item Random forests generally seem to be robust to outliers and noise [\cite{Breiman2001} pages 10, 21; \cite{Wyner2017} pages 4, 12, 17, 20, 32, 35].
\item Boosting tends to perform well and not overfit even when the ensemble consists of many deep trees [\cite{Breiman2001} page 21; \cite{Wyner2017} page 8].
\end{enumerate}

Points 1 and 4 are largely irrelevant to the discussion in the remainder of this paper as we focus exclusively on random forests; we include these points here only in the interest of completeness.  Point 3 has been alluded to in numerous papers throughout the years and has likely been key to the sustained popularity of the random forest procedure.

We take slight issue with the now popular wisdom in the second point, that random forests simply ``are better" than bagging or other similar randomized approaches.  While this does seem to be the case surprisingly often in practice on real-world datasets (see, for example, the recent large-scale comparison from \cite{Fernandez2014} discussed in the introduction) it is certainly not a universal truth and, in our view, is a potentially naive foundation on which to build a theory for explaining their success.  As discussed above, \cite{Breiman2001} does provide some motivation for why a randomized ensemble might sometimes outperform its nonrandomized counterpart in showing that the generalization error of a classifier can be bounded above by a function of base-learner accuracy and correlation.  Breiman stops short, however, of providing any more explicit explanation for the role played by the randomness or in what situations that randomness might be expected to help the most.  \cite{Wyner2017}, on the other hand, seem to largely ignore the role of randomness altogether.  The explanation the authors provide for random forest success would seem to apply equally well to bagging.  In the sections below, we focus our attention heavily on determining when the inclusion of such randomness provides the greatest benefit and provide an explicit characterization of the role it plays.

It is also important to stress that the theories offered by \cite{Breiman2001} and \cite{Wyner2017} pertain only to the classification setting, whereas our focus is primarily on regression.  The interpolation hypothesis put forth by Wyner et al.\ depends on an even stricter setup whereby trees are built to full depth, bootstrapping (or at least subsampling without replacement at a rate of at least $0.5n$) is used to generate resamples, and the number of trees $B$ grows with $n$ at a sufficiently fast rate.  Previous work, however, has repeatedly shown that random forests can still achieve a high degree of predictive accuracy when trees are not built to full depth \citep{Duroux2016}, and/or are constructed via subsampling \citep{Zaman2009,Mentch2016,Wager2018}, and/or when relatively few trees are built \citep{Lopes2019classification}.  

\begin{figure}
	\centering
	\includegraphics[width=0.95\textwidth]{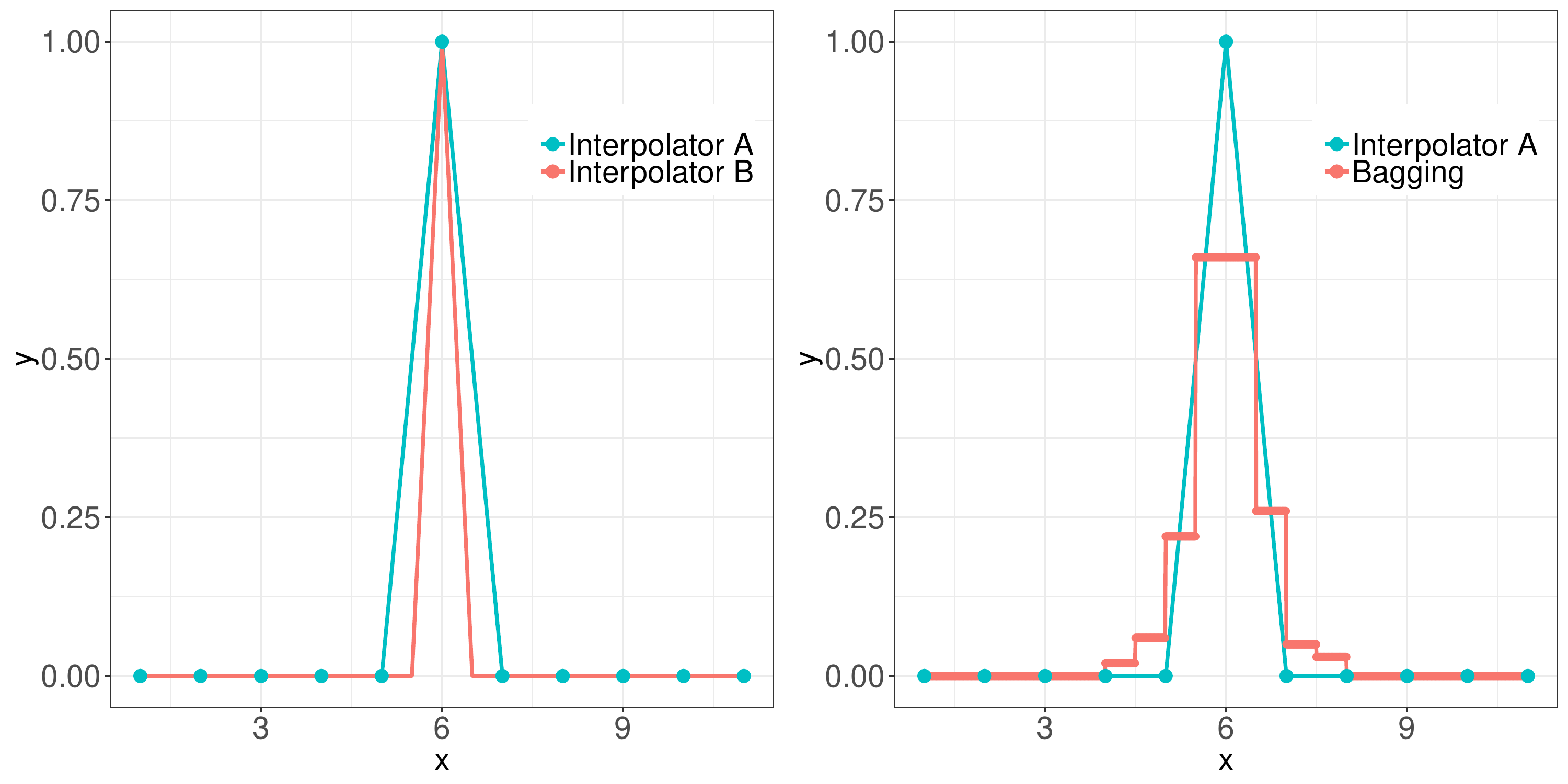}
	\caption{Left: Two interpolating regressors on toy data.  Right:  An interpolating regressor and the predicted regression function resulting from bagging with 100 fully-grown trees.}
	\label{fig:interpolation}
\end{figure}

To see why random forests cannot be considered interpolators in a regression setting, even when individual trees are built to full depth, note that the final regression estimate is taken as the average of predictions across all trees rather than the majority vote.  While it's certainly clear that an average of interpolators is itself an interpolator since for every training point $(\bm{x},y)$
\[
\frac{1}{B} \sum_{i=1}^{B} \hat{f}_i(\bm{x}) = \frac{1}{B} \sum_{i=1}^{B} y = y \, ,
\]
\noindent the bootstrapping mechanism in play with random forests precludes the possibility of interpolating on $\mathcal{D}_n$ with exceedingly high probability.  As noted above, each bootstrap sample will omit, on average, 36.8\% of the original observations and thus individual trees, even if fully grown, will not, in general, fit perfectly on that out-of-sample (out-of-bag) data.  In other words, while a fully-grown tree will necessarily interpolate the observations selected within its corresponding bootstrap sample, it will generally not fit perfectly to all training observations.  For a given point $(\bm{x},y)$ in the training data, random forest regression estimates at $x$ are therefore a weighted average of $y$ and the other response values observed, and hence with exceedingly high probability, the random forest will not interpolate.

Figure \ref{fig:interpolation} demonstrates this effect clearly.  The left panel shows two hand-crafted interpolating functions -- Interpolator A and Interpolator B -- on a simple toy dataset while the right panel shows one of the same interpolators along with the predicted regression function resulting from bagging with 100 regression trees, each grown to full depth.  In this toy example motivated by the examples shown in Section 3.2 of \cite{Wyner2017}, our training data consists of 11 points each of the form $(i,0)$ for $i=1, ..., 11$ except for $i=6$ where we instead have the observation $(6,1)$.  Denote the location of this point by $x^*$.  \cite{Wyner2017} contend that if the observed response is considered ``unusually noisy" at $x^*$, then interpolating estimators can perform well by ``localizing" the effect of this noise as is seen in the left panel of Figure \ref{fig:interpolation}.  Indeed, we can see from this plot that both interpolators still fit the remaining data perfectly despite the presence of the noisy observation.  However, as can be seen in the right-hand panel, whenever we treat this as a regression problem and build trees to full depth, the random forest (in this case, simple bagging since we have only 1 feature) does not interpolate, but instead looks to be attempting to smooth-out the effect of the outlying observation.  

This, however, stands in opposition to the reasoning provided in \cite{Wyner2017}.  Here the authors are highly critical of the traditional statistical notion of signal and noise and seem to take some issue with the general regression setup given in (\ref{eqn:reg}).  To computer scientists, they claim, in many problems ``there is no noise in the classical sense.  Instead there are only complex signals.  There are residuals, but these do not represent irreducible random errors."  But if the widespread empirical success of random forests is really ``not in spite, but because of interpolation" as claimed, then one must believe that real-world data is generally low-noise, a claim argued against firmly by, for example, \cite{Hastie2017}.  Crucially, this means that not only are data largely free of what scientists may think of as classical kinds of noise like measurement error, but also that all sources of variation in the response $Y$ can be explained almost entirely by the available set of predictor variables $X_1, ..., X_p$.  

Perhaps most importantly, if the success of random forests is the result of interpolation and interpolation is beneficial because most real-world datasets have a high signal-to-noise ratio (SNR), then random forests ought not perform well at all on datasets with low SNRs.  Consider again the plot on the left-hand-side of Figure \ref{fig:interpolation}.  If the outlying point $x^*$ at $(6,1)$ is actually the only ``good signal" while the rest of the data are noisy, then the interpolators shown would be isolating signal rather than noise and hence be performing quite poorly.

But this is exactly the opposite of what we see with random forests in practice.  In the following sections, we show repeatedly on both real and synthetic data that relative to procedures like bagging that utilize non-randomized base learners, the benefit of the additional randomness is most apparent in low SNR settings.  In Section \ref{sec:RFdof} we show that the \texttt{mtry} parameter has a direct effect on the degrees of freedom (dof) associated with the procedure, with low values of \texttt{mtry} (i.e.\ more randomness) corresponding to the least flexible model forms with the fewest dof.  Given this, in Section \ref{sec:RFsnr} we go on to show that as expected, the advantage offered by random forests is most dramatic at low SNRs, and that this advantage is eventually lost to bagging at high SNRs.  We also consider the problem from a slightly different perspective and show that the optimal value of \texttt{mtry} is almost perfectly (positively) correlated with the SNR.  We posit that this behavior is due to a regularizing effect caused by the randomness and bolster this claim by demonstrating the same relatively surprising results hold in simpler linear model setups where randomness is injected into a forward selection process.

\section{ Random Forests and Degrees of Freedom}
\label{sec:RFdof}

Recall from the previous section that we assume data of the form $\mathcal{D}_n = \{Z_1, ..., Z_n\}$ where each $Z_i = (\bm{X}_i,Y_i)$, $\bm{X}_i = (X_{1,i}, ..., X_{p,i})$ denotes a vector of $p$ features, $Y \in \mathbb{R}$ denotes the response, and the variables have a general relationship of the form $Y = f(X)+\epsilon$.  Assume further that the errors $\epsilon_1, ..., \epsilon_n$ are uncorrelated with mean 0 and (common) variance $\sigma^{2}$.  Given a particular regression estimate $\hat{f}$ that produces fitted values $\hat{y}_1, ..., \hat{y}_n$, the degrees of freedom \citep{Efron1986,GAMs,Tibshirani2015} of $\hat{f}$ is defined as
\begin{equation}
df(\hat{f}) = \frac{1}{\sigma^2}\sum_{i=1}^n \mathrm{Cov}(\hat{y}_i,y_i).
\label{eqn:dof}
\end{equation}

The degrees of freedom (dof) of a particular estimator is generally understood as a measure of its flexibility; estimators with high dof depend more heavily on the particular values observed in the original data and hence are higher variance.  Understanding the dof associated with various estimation procedures can provide valuable insights into their behavior as well as the situations when they might be expected to perform better or worse relative to a set of alternative methods.  \cite{Tibshirani2015} took an important step in this regard, showing that adaptive procedures like best subset selection (BSS) and forward stepwise selection (FS), even when selecting a model with $k$ terms, had more than $k$ dof because of the increased dependence on the data incurred through the selection aspect.  These additional dof were coined the \emph{search} degrees of freedom.  More recently, \cite{Hastie2017} provided a thorough collection of simulations to demonstrate the predictive advantages of regularized procedures like the lasso and relaxed lasso over BSS and FS, especially in low signal-to-noise ratio (SNR) settings where the SNR is defined as 
\[
\SNR = \frac{\Var(f(x))}{\Var(\epsilon)}.
\]
Much of the work in the following sections was inspired by the approach taken in \cite{Hastie2017} and various portions of the work below follow closely to the setups considered there. 

We begin our work by estimating the dof of random forests under various values of \texttt{mtry}.  In linear model contexts, the dof for different estimators is generally shown by plotting the estimated dof against the average number of nonzero coefficients in the selected models.  In our context with tree-based estimators, we use \texttt{maxnodes} -- the maximum number of terminal nodes that any tree within the forest can have -- as an analogue.  For any given forest with fixed value of \texttt{mtry}, we should expect that as trees are allowed to grow deeper (i.e.\ \texttt{maxnodes} takes larger values), the estimators should become more sensitive to the data and hence incur higher dof.

We consider two general model forms: a linear model
\[
Y = X \beta + \epsilon = X_1 \beta_1 + \cdots + X_p \beta_p + \epsilon
\]
and the model
\[
Y = 0.1e^{4 X_1} +\frac{4}{1+ e^{-20(X_2 -0.5)}} + 3 X_3 +2 X_4 + X_5 + \epsilon \, ,
\]
which we refer to as `MARSadd' as it is additive and first appeared in the work on Multivariate Adaptive Regression Splines (MARS) by \cite{Friedman1991}.  Features in the MARSadd model are sampled independently from $\text{Unif}(0,1)$.  For the linear model, in line with \cite{Hastie2017}, we consider three different settings:
\begin{itemize}
	\item \textbf{Low}: $n=100$, $p=10$, $s=5$
	\item \textbf{Medium}: $n=500$, $p=100$, $s=5$
	\item \textbf{High-10}: $n=100$, $p=1000$, $s=10$
\end{itemize}
\noindent where $n$ is the total (training) sample size, $p$ denotes the total number of features generated, and $s \leq p$ is the number of features with a nonzero coefficient thus considered signal.  Rows of $X \in \mathbb{R}^{n \times p}$ are independently drawn from $N(0,\Sigma)$, where $\Sigma \in \mathbb{R}^{p \times p}$ has entry $(i,j)$ = $\rho^{|i-j|}$.  We take $\rho=0.35$ and set the first $s$ components of $\beta$ equal to 1 with the rest set to 0.  This setup corresponds to the general sampling scheme and `beta-type 2'  setting from \cite{Hastie2017}.  For both models here as well as throughout the majority of the remainder of this paper, we consider sampling the noise as $\epsilon \sim N(0, \sigma^2 I)$ where $\sigma^2$ is chosen to produce a corresponding SNR level $\nu$, so that, for example, in the linear model case, we take 
\[
\sigma^2 = \frac{\beta^T\Sigma\beta}{\nu}.
\]
Finally, in most previous literature, $\texttt{mtry} \leq p \in \mathbb{Z}^{+}$ denotes the number of features eligible for splitting at each node.  Here and throughout the remainder of the paper, we adopt a slightly different (but equivalent) convention by defining \texttt{mtry} as the \emph{proportion} of eligible features so that $\texttt{mtry} \in (0,1]$.  This is purely for readability and ease of interpretation so that readers may more immediately see whether the number of available features is large or small (relative to $p$) regardless of the particular model setup being considered.  

\begin{figure}[t]
	\centering
	\includegraphics[width=0.8\textwidth]{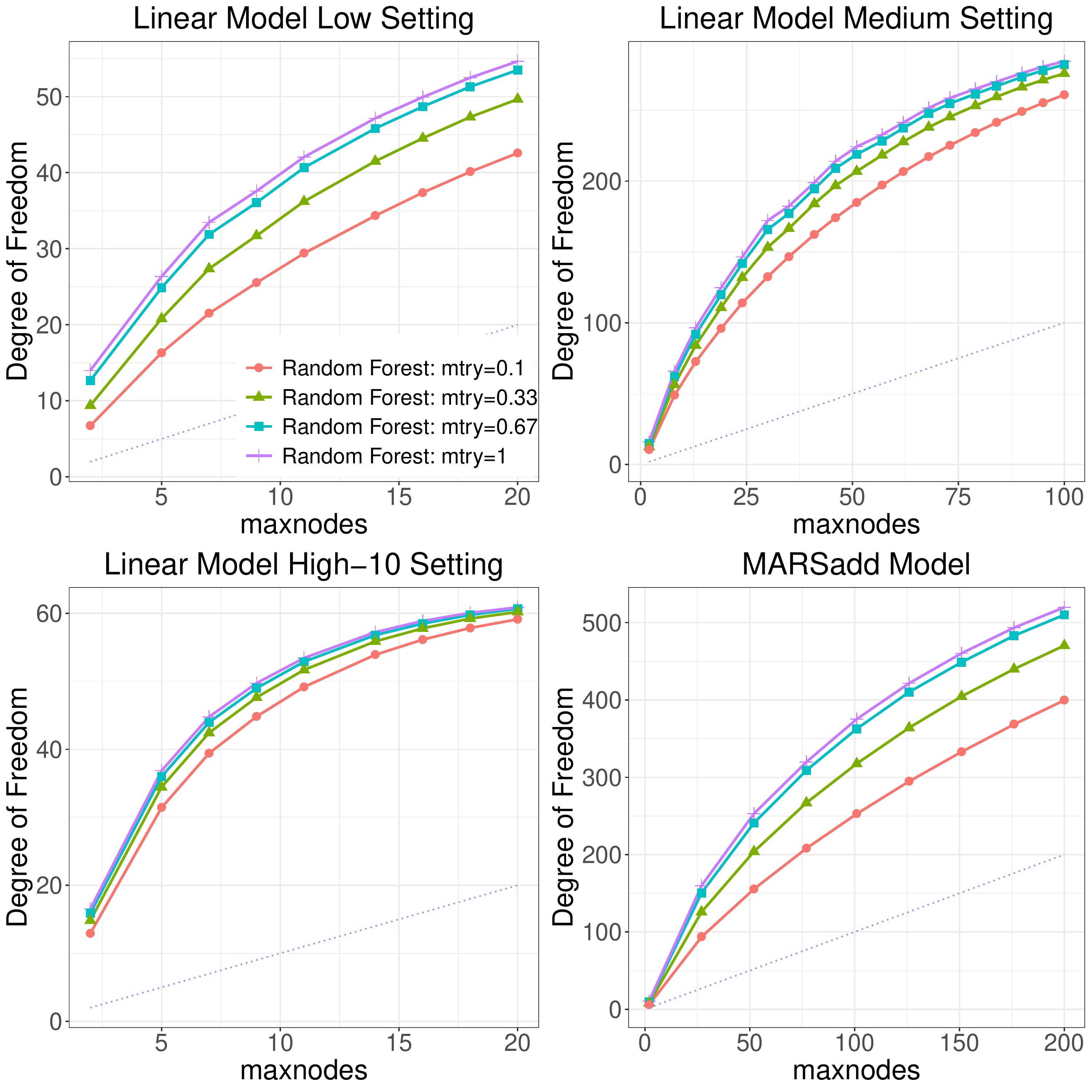}
	\caption{Degrees of freedom for random forests at different levels of \texttt{mtry}.}
	\label{fig:bagrf_df}
\end{figure}

Results are shown in Figure \ref{fig:bagrf_df}. In each of the four model setups we take the SNR to be equal to 3.52 and estimate the dof for random forests with \texttt{mtry} equal to 0.1, 0.33, 0.67, and 1.  Note that when $\mtry = 1$ the model reduces to bagging \citep{Breiman1996} and $\mtry = 0.33$ corresponds to the standard choice of $p/3$ eligible features at each node in regression settings, as is the default in most software.  The forests are constructed using the \texttt{randomForest} package in \texttt{R} \citep{Liaw2002} with the default settings for all arguments except for $\mtry$ and $\maxnodes$.  Each point in each plot in Figure \ref{fig:bagrf_df} corresponds to a Monte Carlo estimate of the dof formula given in (\ref{eqn:dof}) evaluated over 500 trials. 

Several clear patterns are apparent in Figure \ref{fig:bagrf_df}.   First and perhaps most obviously, as conjectured above, in each case we see that the dof increases as $\maxnodes$ increases and trees can be grown to a greater depth.  Each plot also shows the same general concave increasing shape for each forest.  Furthermore, the estimated dof function for each forest lies above the diagonal (shown as a dotted line in each plot), supporting the general notion formalized in \cite{Tibshirani2015} that adaptive procedures like the tree-based models employed here incur additional dof as a result of this search. 

More importantly for our purposes, in each plot in Figure \ref{fig:bagrf_df} we see that at every fixed level of $\maxnodes$, the dof increases with $\mtry$.  In particular, bagging ($\mtry = 1$) always contains more dof than the default implementation for random forest regression ($\mtry = 0.33$).  Finally, we note that the patterns seen in these plots also hold for numerous other regression functions and SNRs that were experimented with; Figure \ref{fig:bagrf_df_app} in Appendix \ref{app:plots} shows the results of the same experiments above carried out at a much lower SNR of 0.09 and the findings are nearly identical.

\section{Random Forest Performance vs Signal-to-Noise Ratio}
\label{sec:RFsnr}

The empirical results in the preceding section suggest that the $\mtry$ parameter in random forests is directly tied to its dof with larger values resulting in estimators with higher dof and more flexibility.  Based on this intuition and the results for linear estimators provided in \cite{Hastie2017}, we should therefore expect that random forests with smaller values of $\mtry$ to perform well in noisy settings while bagging should perform best -- potentially even better than random forests -- at high SNRs.  We now investigate this more formally.

\subsection{Relative Performance on Synthetic Data}
We begin by comparing the relative performance of random forests and bagging on simulated data across a range of plausible SNRs.  In addition to the linear model setup described in the previous section, we now include the additional regression function
\[
	Y=10\sin(\pi X_1 X_2) + 20(X_3- 0.05)^2 +10X_4+5X_5 + \epsilon
\]
which we refer to as `MARS' because like the additive model used previously, it first appeared in the MARS paper \citep{Friedman1991}, though note that unlike the previous model, it contains explicit interactions between the features.  This particular MARS model has proven popular in random forest publications in recent years, appearing, for example, in \cite{Biau2012} and \cite{Mentch2016}.  In the simulation setups described below, we consider the medium setting for the linear model ($n = 500$, $p = 100$, $s = 5$) and for the MARS model, we take $p = s = 5$ and consider sample sizes of $n = 200, 500, $ and 10000.  Features for the linear model are generated in the same fashion as above and those in the MARS model are sampled independently from $\text{Unif}(0,1)$.

As in the previous section, the variance $\sigma^2$ of the noise term is chosen so as to induce particular SNRs.  Here, following \cite{Hastie2017}, we consider 10 SNR values $\nu = 0.05, 0.09, 0.14, ..., 6.00$ equally spaced between 0.05 and 6 on the log scale.  Forests are again constructed using the \texttt{randomForest} package with the default settings except in the case of bagging where the default value of $\mtry$ is changed so that all features are available at each split.

Here, in comparing the performance of ``random forests" against ``bagging", we stress that we are merely assessing the difference in predictive accuracies between forests constructed with $\mtry = 0.33$ (traditional random forests) versus those built with $\mtry = 1$ (bagging).  To compare the relative performance for a fixed model setup with fixed SNR, we first generate a training dataset and then evaluate the mean squared error (MSE) for both bagging and random forests on an independently generated test dataset consisting of 1000 observations.  This entire process is then repeated 500 times for each setting and we record the average difference in accuracy ($\text{Error(Bagg)} - \text{Error(RF)} $) across these repititions. 

\begin{figure}[t]
	\centering
	\includegraphics[width=0.8\textwidth]{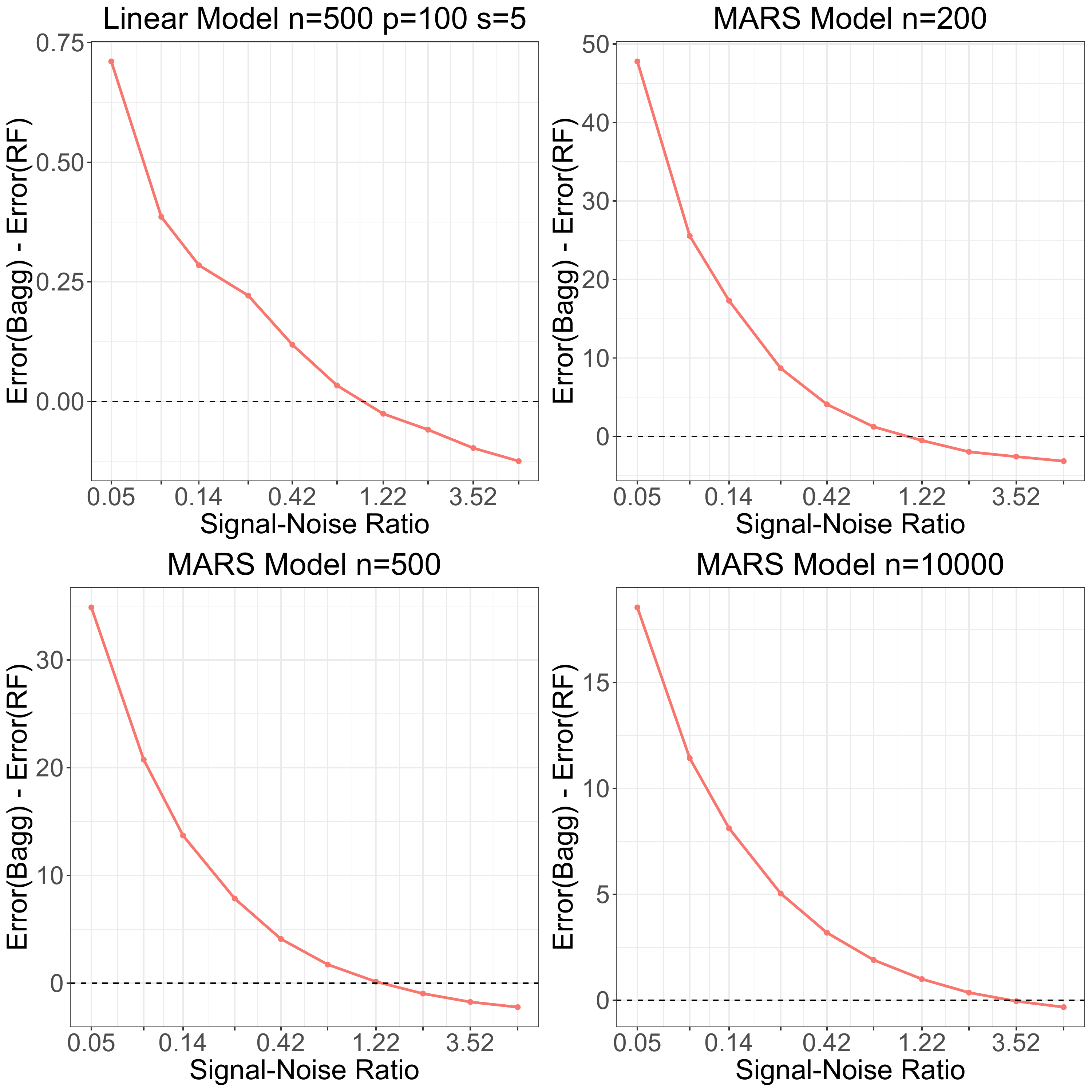}
	\caption{Differences in test errors between bagging and random forests.  Positive values indicate better performance by random forests.}
	\label{fig:bagrf_snr_syn}
\end{figure}

Results are shown in Figure \ref{fig:bagrf_snr_syn}.  Each plot shows the average difference in test errors versus the SNR; note that positive differences indicate that random forests ($\mtry = 0.33$) are outperforming bagging ($\mtry = 1$).  In each of the four regression setups shown in Figure \ref{fig:bagrf_snr_syn}, the improvement in accuracy seen with random forests decreases as the SNR increases.  For large SNR values, bagging eventually begins to outperform the traditional random forests.  Note also that for the MARS function, random forests seem to retain their relative improvement longer (i.e.\ for larger SNRs) with larger training samples.  Given these results, the conventional wisdom that random forests simply ``are better" than bagging seems largely unfounded; rather, the optimal value of $\mtry$ seems to be a function of the SNR.  

\subsection{Optimal \texttt{mtry} vs SNR}
\label{sec:optmtry}
The results above indicate that random forests ($\mtry = 0.33$) generally seem to outperform bagging ($\mtry = 1$) unless the SNR is large.  We now reverse the direction of this investigation and estimate the optimal value of $\mtry$ across various SNR levels.

Here, as above, we consider both the MARS model and a linear model.  In the same fashion as in previous simulations, the errors are sampled from a $N(0, \sigma^2)$ where $\sigma^2$ is chosen to produce a particular SNR and we consider the same 10 SNR values as above.  For the MARS model, we take $p = s = 5$ and generate features independently from $\text{Unif}(0,1)$ and for the linear model, we take $p = 20$ and $s = 10$ with features drawn from $N_p(0,\Sigma)$ where the $(i,j)$ entry of $\Sigma$ is given by $\rho^{|i-j|}$ with $\rho=0.35$.  The first $s = 10$ coefficients in the linear model are set equal to 1 with the rest set equal to 0.

\begin{figure}[t]
	\centering
	\includegraphics[width=0.48\textwidth]{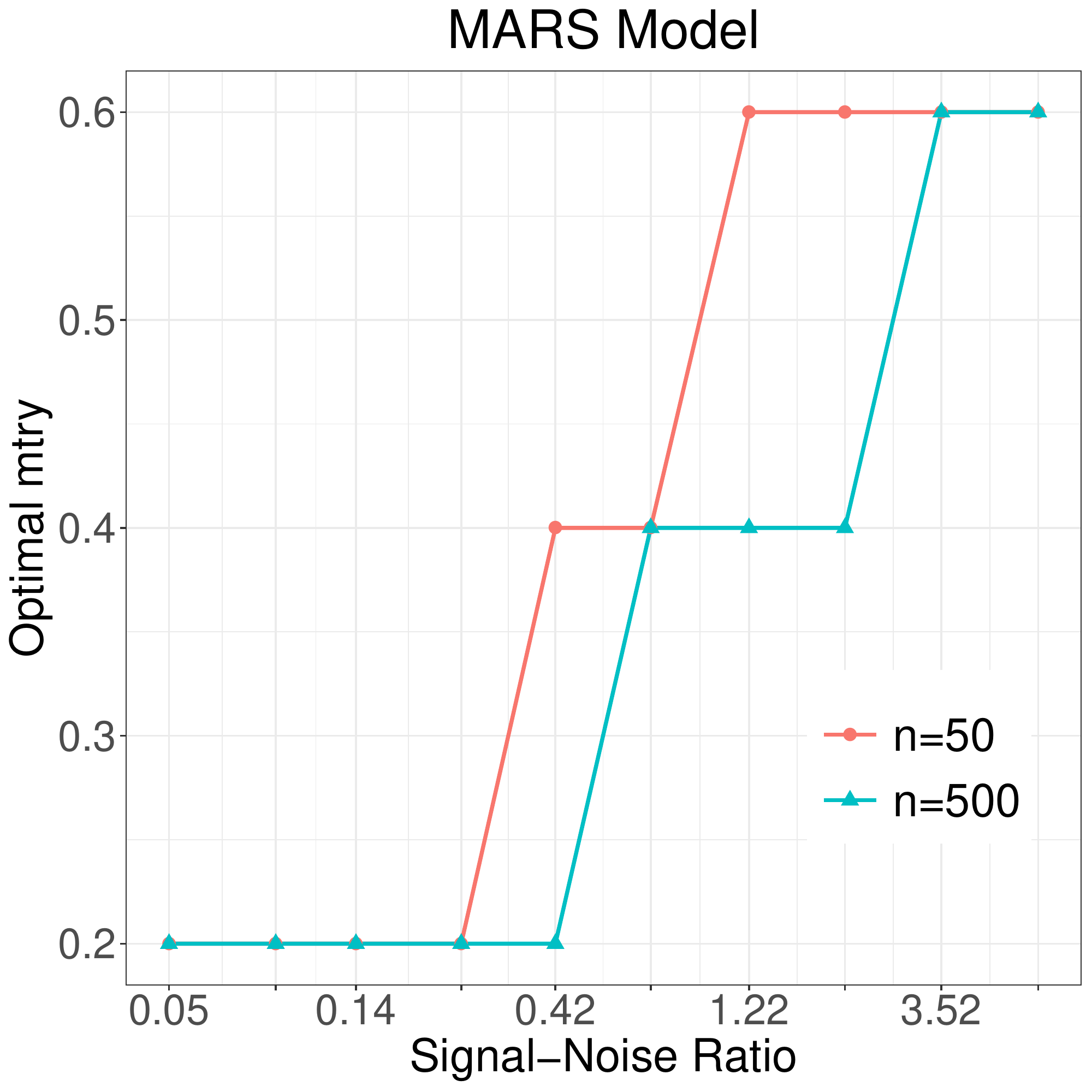}
	\includegraphics[width=0.48\textwidth]{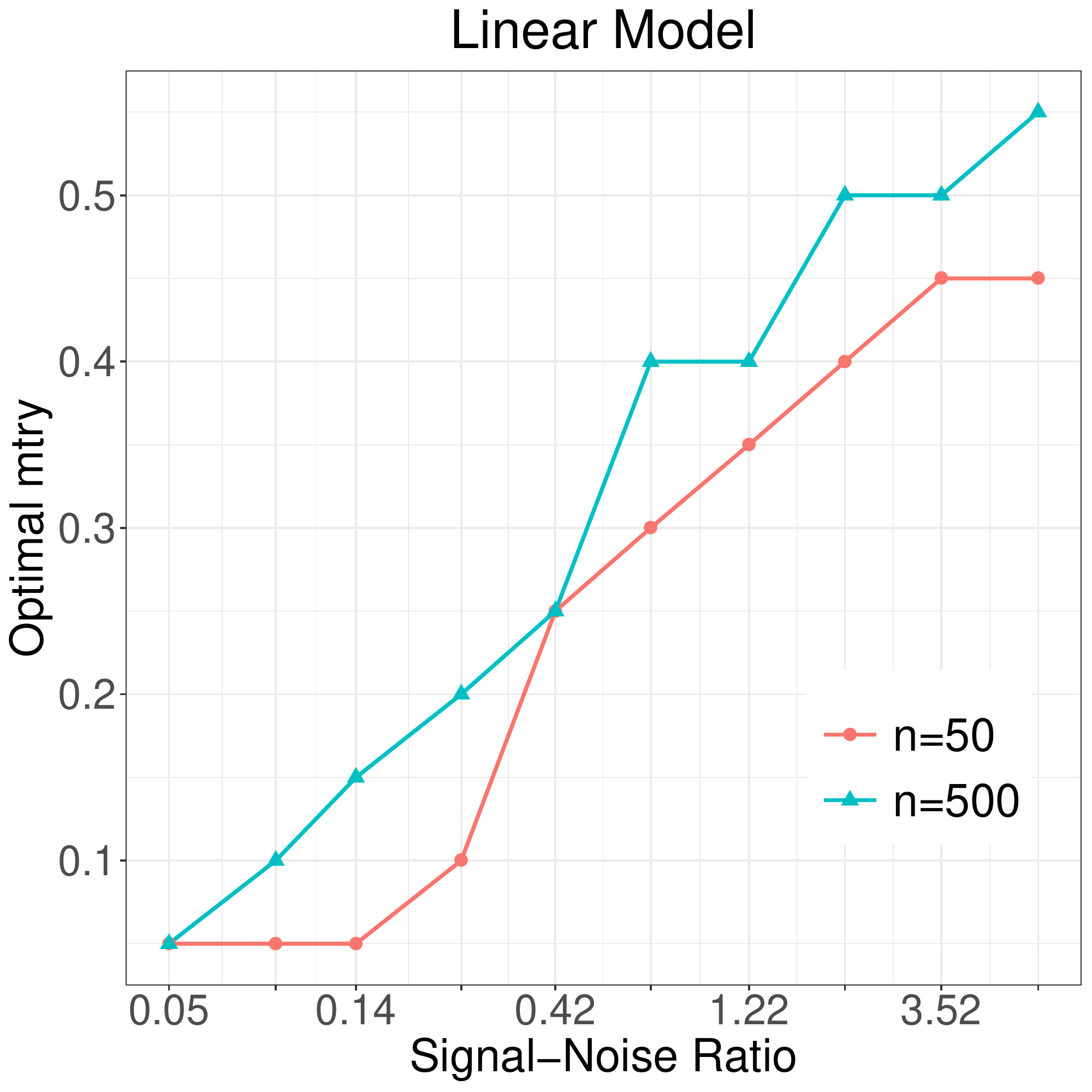}
	\caption{ Optimal value of $\mtry$ vs SNR for the MARS and linear model.}
	\label{fig:optmtry}
\end{figure}

For both models, we consider (training) sample sizes of both $n = 50$ and $n = 500$ and generate an independent test set of the same size.  We construct forests using all possible values of $\mtry$ with the remaining options at the default settings in \texttt{randomForest}.  The entire process is repeated 500 times and the $\mtry$ value corresponding to the forest with the lowest average test error for each setting is selected.  The results are shown in Figure \ref{fig:optmtry}.  As expected, corroborating the findings above, the optimal value of $\mtry$ increases with the SNR and the same general pattern emerges for both models and sample sizes.  Figure \ref{fig:optmtrymean} in Appendix \ref{app:plots} shows a slightly different calculation where the optimal $\mtry$ value on each of the 500 iterations is determined and the overall mean is then calculated.  Here too we see exactly the same general pattern in that as the SNR increases, so does the optimal value of $\mtry$.

\subsection{Relative Performance on Real Data}
The work above presents strong empirical evidence that the relative improvement in predictive accuracy seen with random forests is largest at low SNRs and more generally, that the optimal value of $\mtry$ appears to be a direct (increasing) function of the SNR.  These results, however, pertain only to those particular simulation settings that some may argue are highly idealized.  Real-world data may contain far more complex relationships and thus we now explore whether the same general findings above also appear in more natural contexts.  

\begin{table}[t]
	\centering
	\begin{tabular}{lcc}
	\hline
	Dataset	&	$p$	&	 $n$		\\
	\hline
	Abalone Age [\texttt{abalone}] \citep{UCIabalone}	&	8			&	4177		\\
	Bike Sharing [\texttt{bike}] \citep{UCIbike}	&	11			&	731		\\
	Bioston Housing [\texttt{boston}]	\citep{UCIboston} &	13			&	506		\\
	Concrete Compressive Strength [\texttt{concrete}] \citep{UCIconcrete}	&	8			&	1030		\\
	CPU Performance [\texttt{cpu}] \citep{UCIcpu}	&	7			&	209		\\
	Conventional and Social Movie [\texttt{csm}] \citep{UCIcsm}	&	10			&	187		\\
	Facebook Metrics [\texttt{fb}] \citep{UCIfb}	&	7			&	499		\\
	Parkinsons Telemonitoring [\texttt{parkinsons}] \citep{UCIparkinsons}	&	20			&	5875		\\
	Servo System [\texttt{servo}] \citep{UCIservo}	&	4			&	167		\\
	Solar Flare [\texttt{solar}] \citep{UCIflare} &	10			&	1066		\\
	Aquatic Toxicity [\texttt{AquaticTox}] \citep{he2005assessing}	&	468	&	322	\\
	Molecular Descriptor Influencing Melting Point [\texttt{mtp2}] \citep{bergstrom2003molecular}	&	1142			&	274		\\
	Weighted Holistic Invariant Molecular Descriptor [\texttt{pah}] \citep{todeschini1995weighted}	&	112			&	80 		\\
	Adrenergic Blocking Potencies [\texttt{phen}] \citep{cammarata1972interrelation}	&	110			&	22			\\
	PDGFR Inhibitor [\texttt{pdgfr}] \citep{guha2004development}	&	320			&	79			\\
	\hline
	\end{tabular}
	\caption{Summary of real-world data utilized. For datasets where no reference was specified, a reference to early work utilizing the data is given.}
	\label{tab:uci}
\end{table}

To investigate this, we utilize 10 datasets intended for regression from the UCI Machine Learning Repository \citep{uci}.  Because most of these datasets are low-dimensional, five additional high-dimensional datasets were also included, four of which were downloaded from \texttt{openml.org} \citep{OpenML2013} with the other (\texttt{AquaticTox}) taken from the \texttt{R} package \texttt{QSARdata}.  Summaries of these datasets are provided in Table \ref{tab:uci}.  For datasets containing missing values (\texttt{csm} and \texttt{fb}), the corresponding rows of data were removed.  Here we do not know the true SNR and thus to compare the relative performance of bagging and random forests, we inject additional random noise $\epsilon$ into the response, where each $\epsilon \sim N(0,\sigma^2)$ and $\sigma^2$ is chosen as some proportion $\alpha$ of the variance of the original response variable.  We consider $\alpha = 0, 0.01, 0.05, 0.1, 0.25 \text{ and } 0.5$ where $\alpha=0$ corresponds to the case where no additional noise is added and performance is thus compared on the original data.  To compare performance, we measure the relative test error (RTE)
\begin{equation}
\text{RTE } = \frac{   \widehat{Err}(\text{Bagg}) - \widehat{Err}(\text{RF})}{\hat{\sigma}_{y}^2} \times 100\%
\label{eqn:RTE}
\end{equation}
\noindent where $\widehat{Err}(\text{Bagg})$ and $\widehat{Err}(\text{RF})$ denote the 10-fold cross-validation error on bagging and random forests, respectively, and $\hat{\sigma}_{y}^{2}$ is the empirical variance of the original response.  For each setting on each dataset, the process of adding additional random noise is replicated 500 times and the results are averaged.  Once again, forests are constructed using the \texttt{randomForest} package with the default settings except for fixing $\mtry = 1$ for bagging and $\mtry = 0.33$ for random forests.

\begin{figure}
	\centering
	\includegraphics[width=0.475\textwidth]{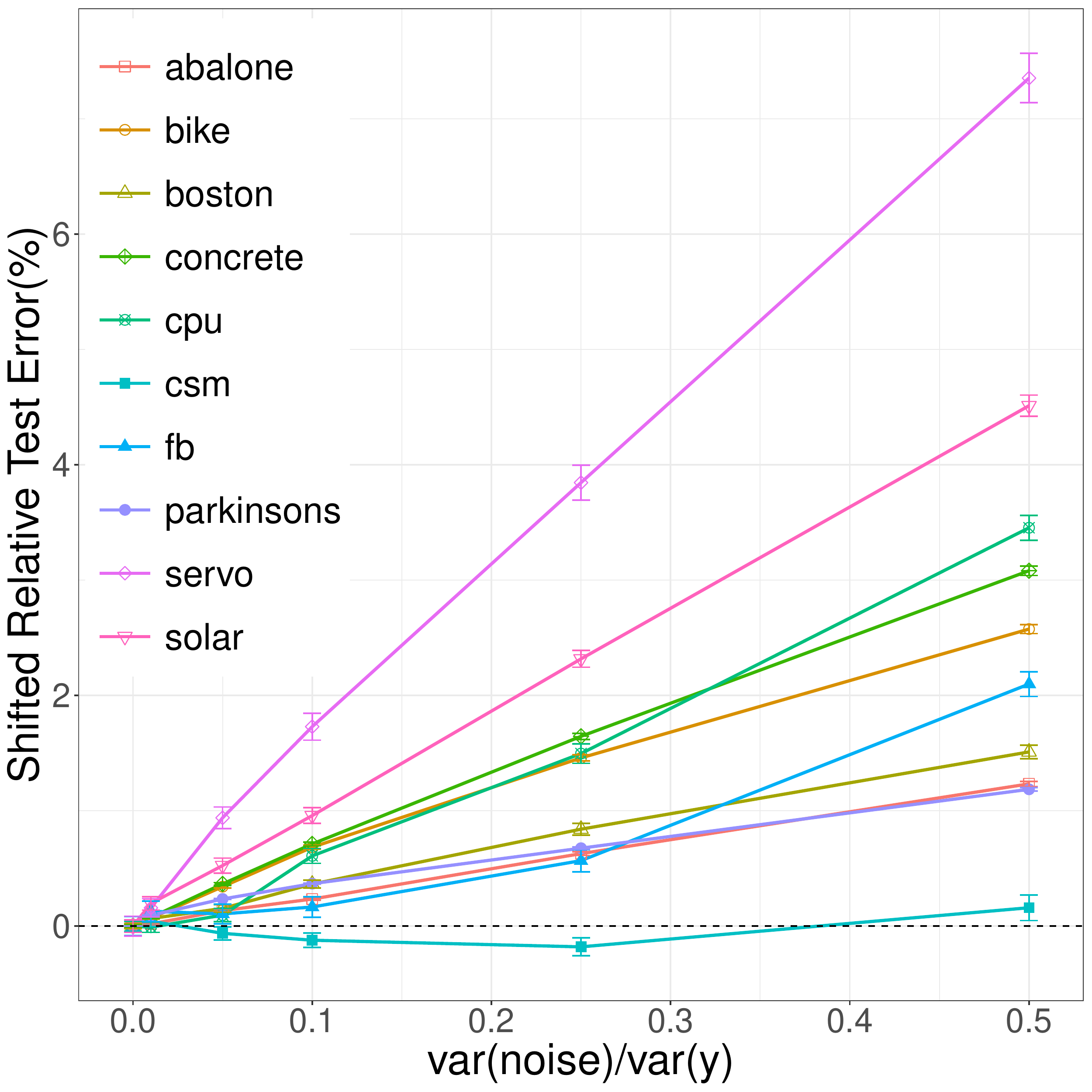}
	\includegraphics[width=0.475\textwidth]{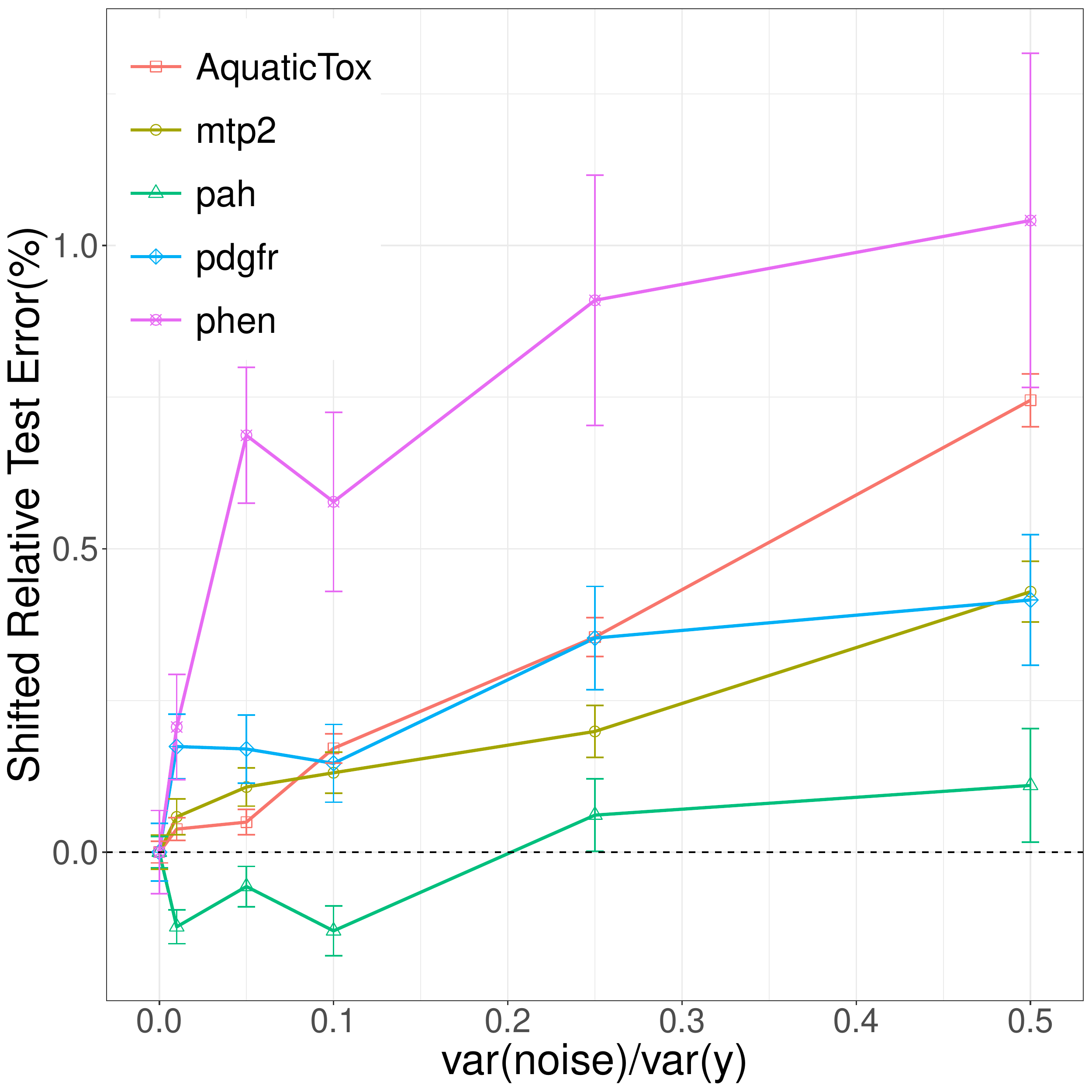}
	\caption{Shifted RTE on real data where additional noise is added.  The left plot shows results on low-dimensional datasets taken from the UCI repository; the right plot shows results on high-dimensional datasets.}
	\label{fig:bagrf_snr_real}
\end{figure}

Results are shown in Figure \ref{fig:bagrf_snr_real} with low-dimensional datasets shown in the left plot and high-dimensional datasets shown on the right.  Note that to aid in presentation, these display the \emph{shifted} RTE rather than the raw calculation in (\ref{eqn:RTE}).  For a given proportion of additional noise $\alpha$, let RTE($\alpha$) denote the corresponding relative test error.  The shifted RTE at noise level $\alpha$ is then defined as RTE($\alpha$) - RTE(0).  This ensures that the relative error for each dataset begins at the origin thereby allowing us to present all results in a single easy-to-interpret plot.  Error bars correspond to $\pm1$ standard deviation across 500 trials.

In both plots in Figure \ref{fig:bagrf_snr_real}, the same general pattern appears as has been seen in the subsections above:  as we increase the amount of additional noise inserted into the models, the relative improvement in predictive accuracy seen with random forests becomes more pronounced.  The only slight exceptions are seen on the low-dimensional \texttt{csm} dataset and the high-dimensional \texttt{pah} dataset were the relative error appears to decrease by a small amount before eventually coming back up when large amounts of noise are added.  It is not clear why the initial temporary drops occur in these two datasets, though it is worth noting that even the largest magnitudes of drops are quite small at approximately 0.2\% and 0.13\% in the \texttt{csm} and \texttt{pah} datasets, respectively.

\section{Randomized Forward Selection}
\label{sec:RFS}

The results from the previous sections suggest that the optimal value of $\mtry$ for random forests is highly data-dependent, with smaller values preferred in noisy settings and bagging ($\mtry = 1$) being preferred when very little noise is present.  These findings are much in line with the classic understanding of random forests as a means by which the variance of the ensemble is reduced as a by-product of reducing the correlation between trees.  The benefits of this variance reduction are most apparent at low SNRs.  

In our view, however, this remains only a partial explanation.  While randomizing the collection of features eligible for splitting at each node is one way to reduce between-tree correlation, it is certainly not the only means by which this can be accomplished.  \cite{Breiman2001} experimented with alternative implementations finding that even very naive approaches like adding random noise to the outputs of each tree could sometimes be beneficial.  But as discussed in the opening sections, \cite{Breiman2001}, like many others after, also noted that the particular approach where features are randomly precluded from splitting at each node seemed to produce substantially more accurate predictions than other strategies for reducing between-tree correlation.  Why this was the case, however, was not clear and has remained a subject of speculation in the nearly two decades following the inception of the procedure.

As already briefly mentioned above, \cite{Hastie2017} recently provided an extended comparison of several variable selection procedures including the lasso \citep{Tibshirani1996,Chen2001}, relaxed lasso \citep{Meinshausen2007}, forward stepwise selection (FS), and best subset selection (BSS).  The relaxed lasso estimator utilized in \cite{Hastie2017} takes the form
\[
\hat{\beta}_{\text{relax}}(\lambda,\gamma) = \gamma \hat{\beta}_{\text{lasso}}(\lambda) + (1-\gamma) \hat{\beta}_{\text{LS} | \text{lasso}}(\lambda)
\]
where $\hat{\beta}_{\text{LS} | \text{lasso}}(\lambda)$ denotes the vector of coefficient estimates obtained via least squares (LS) when computed on only those variables selected via the lasso and filled in with 0 for variables not selected.  Perhaps the most striking takeaway from their study is that in low-dimensional settings where $n > p$, the more aggressive procedures (FS and BSS) with higher dof are generally not competitive with a regularized approach like the lasso at low SNRs but eventually produce more accurate predictions when the SNR becomes large.  Relaxed lasso, taking the weighted average of lasso and LS-after-lasso estimates, possesses the ability to effectively trade-off the amount of regularization needed depending on the SNR and always seems to perform well.    

For these kinds of estimators whose inner-workings are better understood, the reasoning behind the results observed is relatively straightforward.  In low SNR settings, procedures like the lasso and relaxed lasso that explicitly regularize the problem can prevent overfitting to the noise by applying shrinkage to the coefficient estimates of the selected features.  Given that we see the same general pattern here -- random forests ($\mtry = 0.33$) outperforming bagging ($\mtry = 1$) except at high SNRs -- it is reasonable to suspect that the additional randomness in random forests is playing a similar regularization role.  Indeed, by randomly not allowing certain features to be split, random forests may be seen as effectively shrinking the potential influence of features, with the amount of shrinking being proportional to the amount of additional randomness added (with smaller values of $\mtry$ inducing more randomness).  Speculation to this effect is described in \cite{esl}.

Importantly however, if this is the kind of underlying effect that allows random forests to perform well in practice, such an effect should not be limited to tree-based ensembles.  Indeed, if regression trees are seen as merely a complex form of forward selection, then if we were to create bagged and randomized versions of a standard forward selection procedure -- analogues to the traditional tree-based versions of bagging and random forests -- we should expect to see the same general patterns of relative improvement.  In the following sections, we propose two such ensemble-ized extensions of forward selection and confirm that not only do similar patterns emerge, but that these new procedures exhibit surprisingly strong performance relative to alternative procedures.

\subsection{Degrees of Freedom for Randomized Forward Selection}
\label{sec:dofRandFS}

\begin{algorithm}[t]
	\caption{Randomized Forward Selection (\texttt{RandFS})}
	\label{alg:RandFS}
	\begin{algorithmic}
	\Procedure{RandFS}{$\mathcal{D}_n, B, d, \texttt{mtry}$}
	\For{$b = 1,\dots,B $}  
			\State Draw bootstrap sample $\mathcal{D}^{(b)} =  \{(\bm{X}_i^{(b)},Y_i^{(b)}) \}_{i=1}^{n}$ from original data $\mathcal{D}_n$
			
			\State Initialize empty active set $A_0 = \{0\} $
			\For{$k \in 1 \dots d$}		
				\State Select subset of $\mtry \times p$ features uniformly at random, denoted $F_k$
				\State Select $j_k = \argmin_{j  \in F_k}  \norm{Y^{(b)} - P_{A_{k-1}\cup \{j \}} Y^{(b)}}^2_2$

				\State Update active set $A_{k}=A_{k-1} \cup \{j_k\}$
				\State Update coefficient estimates $\hat{\beta}^{(b)}$ as 
					\[
					\hat{\beta}^{(b)}_{A_k} = \argmin_{\beta} \norm{Y^{(b)} - \bm{X}_{A_k}^{(b)} \beta}^2 , \quad \hat{\beta}^{(b)}_{A_k^c} = 0
					\]
			\EndFor
		\EndFor
		\State Compute final coefficient estimates $\hat{\beta} = \frac{1}{B} \sum_{b=1}^B \hat{\beta}^{(b)}$
		\State Compute predictions $\hat{Y} = X\hat{\beta}$
	\EndProcedure
	\end{algorithmic}
\end{algorithm}

We begin by formalizing the notion of randomized forward selection (RandFS), which can
be seen as a random forest analogue to traditional forward stepwise selection (FS). For any subset of the feature indices $\mathcal{S}$, define $\bm{X}_{\mathcal{S}}$ as the matrix of feature values whose index is in $\mathcal{S}$ and $P_{\mathcal{S}}$ as the projection matrix onto the column span of $\bm{X}_{\mathcal{S}}$. To carry out randomized forward selection (\texttt{RandFS}), we begin by drawing $B$ bootstrap samples from the original data and performing forward selection on each. However, like random forests, at each step, only a random subset of remaining features are eligible to be included in the model.  The process continues until the desired model size (depth) $d$ is obtained and the final predictions are taken as an average over the predictions generated by each individual model.  When $\mtry = 1$ so that all features are eligible at each step, we refer to the procedure as \emph{bagged} forward selection (\texttt{BaggFS}). A summary of the procedure is given in Algorithm \ref{alg:RandFS}.

We begin by estimating the dof for RandFS as well as for FS, lasso, and relaxed lasso.  Here we follow the same initial setup utilized in \cite{Hastie2017} where we assume a linear model $Y = X\beta +\epsilon$ with $n=70$, $p=30$, and $s=5$.  Rows of $X$ are sampled independently from $N_p(0,\Sigma)$, where the $(i,j)^{th}$ entry of $\Sigma$ takes the form $\rho^{|i-j|}$ with $\rho=0.35$ and errors are sampled independently from $N(0, \sigma^2)$ with $\sigma^2$ chosen to satisfy a particular SNR, in this case 0.7.  The first $s$ components of $\beta$ are set equal to 1 with the rest equal to 0.  

\begin{figure}[t]
\centering
\includegraphics[width=0.48\textwidth]{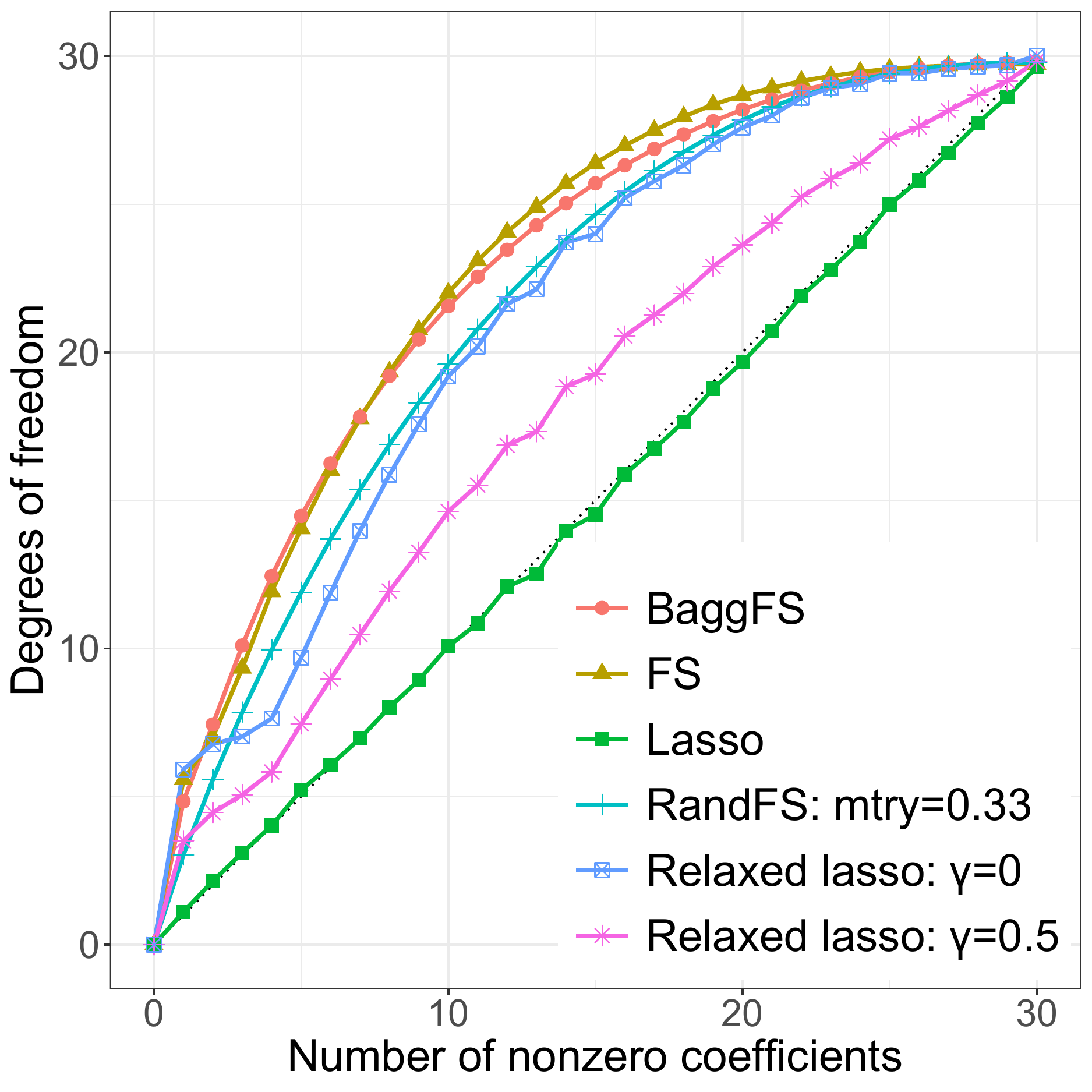}
\includegraphics[width=0.48\textwidth]{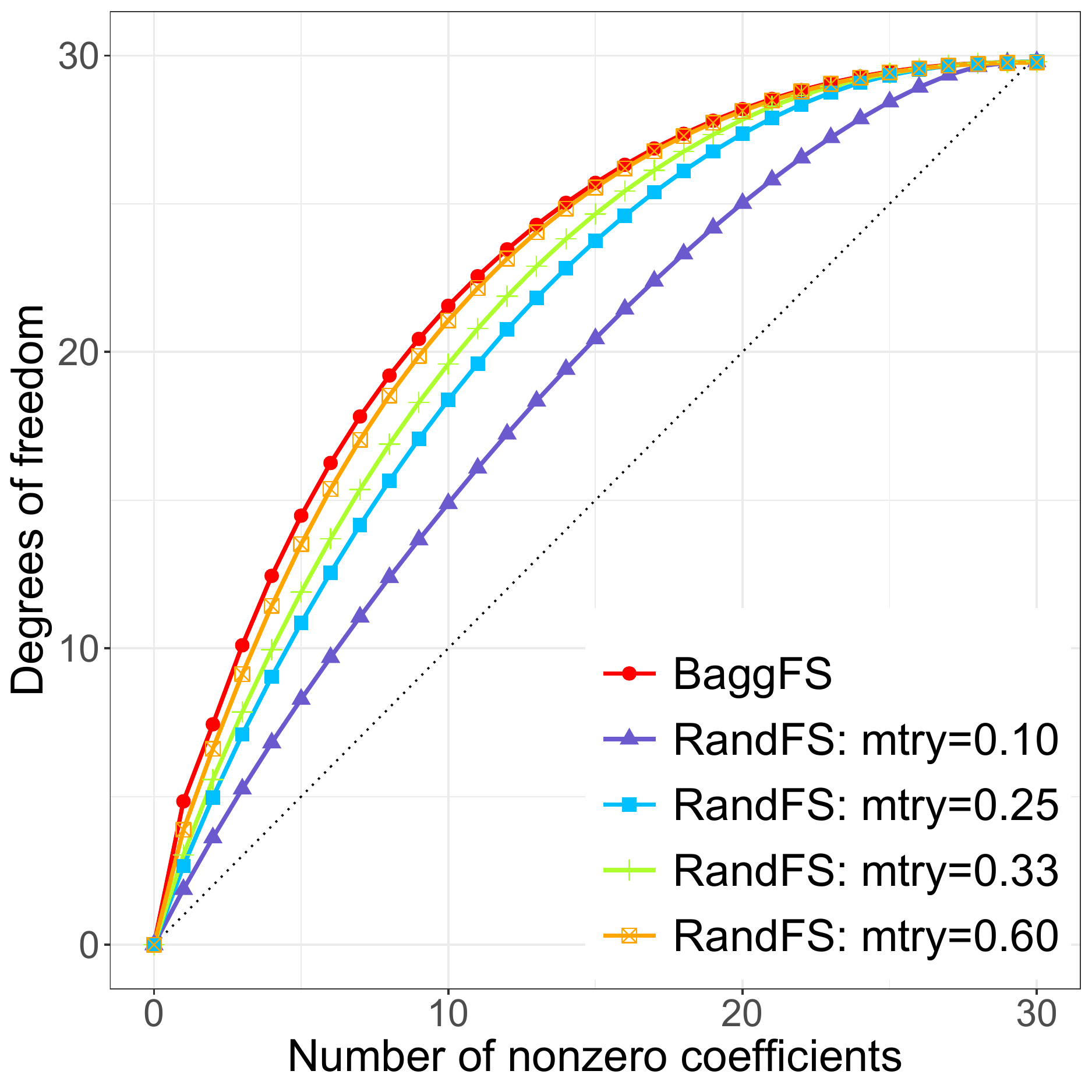}
\caption{Estimated dof for forward selection (FS), bagged forward selection (BaggFS), randomized forward selection (RandFS), lasso, and relaxed lasso.}
\label{fig:randfwd_df_1}
\end{figure}

The plots in Figure \ref{fig:randfwd_df_1} show the estimated dof for the various methods against the number of nonzero coefficient estimates produced.  Each point in each plot corresponds to a Monte Carlo estimate of the dof formula given in (\ref{eqn:dof}) evaluated over 500 iterations.  As expected, the plot on the right shows quite clearly that as with random forests, larger values of $\mtry$ produce RandFS procedures with more dof.  More surprising is the relative relationship of the RandFS models to the more classical procedures.  The plot on the left shows that the dof for BaggFS is almost identical to that of standard FS.  The standard $\mtry$ value of 0.33 produces a RandFS model with dof very similar to that of the relaxed lasso with $\gamma = 0$, corresponding to least squares after lasso.  Smaller values of $\mtry$ appear to offer even more regularization, producing dof similar to relaxed lasso with larger $\gamma$ values, corresponding to an estimate where more weight is put on the original lasso coefficient estimates.  

\subsection{Relative Performance of Randomized Forward Selection}
\label{sec:RandFSSNR}
Building on the intuition from previous sections as well as that provided in \cite{Hastie2017}, the dof results in Figure \ref{fig:randfwd_df_1} suggest that we should expect to see RandFS models with small values of $\mtry$ have a potential advantage in predictive accuracy relative to FS and BaggFS at low SNRs.  We now compare the performance of RandFS relative to BaggFS and the more classical procedures.

Here we consider several linear model setups taken directly from \cite{Hastie2017} and similar to those considered in Section \ref{sec:RFdof}.  We consider four settings:
\begin{itemize}
	\item \textbf{Low}: $n=100$, $p=10$, $s=5$
	\item \textbf{Medium}: $n=500$, $p=100$, $s=5$
	\item \textbf{High-5}: $n=50$, $p=1000$, $s=5$
	\item \textbf{High-10}: $n=100$, $p=1000$, $s=10$.
\end{itemize}
\noindent As above, rows of $X$ are independently drawn from $N(0,\Sigma)$, where $\Sigma \in \mathbb{R}^{p \times p}$ has entry $(i,j)$ = $\rho^{|i-j|}$ with $\rho=0.35$ and where we set the first $s$ components of $\beta$ equal to 1 with the rest set to 0, corresponding to the beta-type 2 setting in \cite{Hastie2017}.  Noise is once again sampled from $N(0, \sigma^2 I)$ where $\sigma^2$ is chosen to produce a corresponding SNR level $\nu$ and we consider the same 10 values $\nu = 0.05, 0.09, 0.14, ..., 6.00$ utilized above.

Tuning of the FS, lasso, and relaxed lasso procedures is done in exactly the same fashion as in \cite{Hastie2017}.  In all cases, tuning parameters are optimized on an independent validation set of size $n$.  For the low setting, the lasso shrinkage parameter $\lambda$ follows the default \texttt{glmnet} settings being tuned across 50 values ranging from small to large fractions of $\lambda_{\max} = \parallel X^T Y \parallel_{\infty}$.  For relaxed lasso, $\lambda$ is tuned across the same 50 values and the $\gamma$ parameter that weights the average of the lasso and LS-after-lasso estimates is chosen from 10 equally spaced values between 0 and 1.  The depth of the models in FS, BaggFS, and RandFS are tuned across $d = 0, 1, 2, ..., 10$.  Note that for BaggFS and RandFS, a selected depth of $d$ means that each individual model is built to a depth of $d$ and the final average is then taken; since different individual models will generally select different features, the final averaged model will generally contain more than $d$ features.  In addition to considering the default RandFS ($\mtry = 0.33$) and BaggFS ($\mtry = 1$), we also consider tuning the $\mtry$ in RandFS across 10 equally spaced values between 0.1 and 1.  In the medium, high-5, and high-10 settings, the $\lambda$ parameter in lasso and relaxed lasso is tuned across 100 values rather than 50 and model depths for FS, BaggFS, and RandFS are tuned across $d = 0, 1, 2, ..., 50$; all other setups remain the same.  

\begin{figure}[t]
\centering
\includegraphics[width=0.8\textwidth]{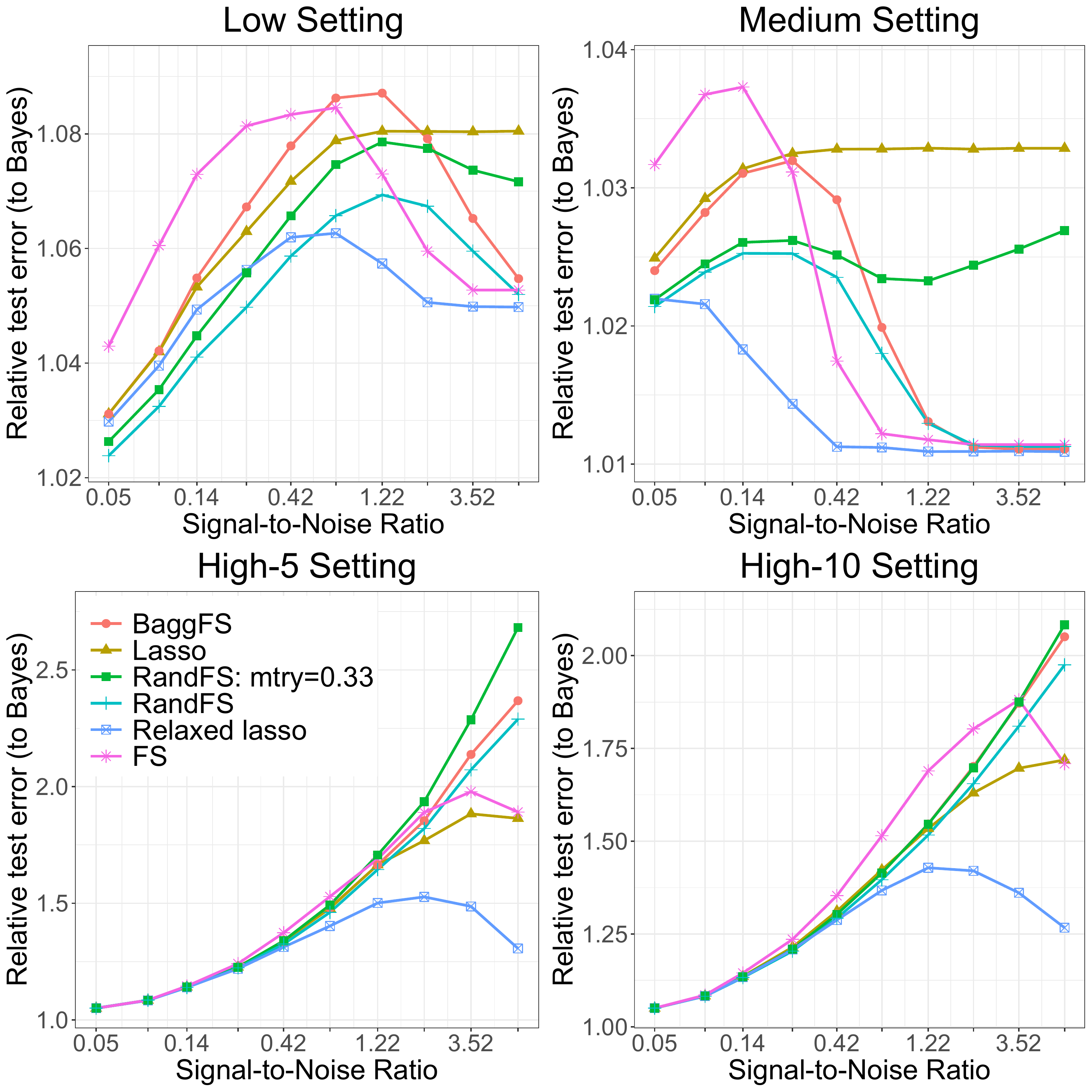}
\caption{Performance of FS, BaggFS, RandFS, lasso and relaxed lasso across SNR levels for linear models in the low, medium, high-5, and high-10 settings. }
\label{fig:randfwd_performance}
\end{figure}

To measure performance, we again follow the lead of \cite{Hastie2017} and calculate the test error relative to the Bayes error rate.  Specifically, given a test point $z_0 = (x_0,y_0)$ with $y_0 = x_0^T \beta + \epsilon_0$ and $\epsilon_0 \sim N(0,\sigma^2)$, the relative test error (RTE) to Bayes of a regression estimate $\hat{\beta}$ is given by
\[
\text{RTE}(\hat{\beta}) = \frac{\mathbb{E}(y_0 - x_0^T\hat{\beta})^2}{\sigma^2} 
			= 	\frac{(\hat{\beta}-\beta)^T\Sigma(\hat{\beta}-\beta)+\sigma^2}{\sigma^2} .
\]

Results are shown in Figure \ref{fig:randfwd_performance}; the same plots with error bars corresponding to $\pm 1$ standard deviation are shown in Appendix \ref{app:plots}.  Each point in each plot corresponds to an average over 100 replications.  As expected, the explicit regularizers (lasso and relaxed lasso) perform well in the high-dimensional settings and at low SNRs.  In the high-dimensional settings in particular, most methods perform similarly at low SNRs but for larger SNRs, relaxed lasso begins to perform substantially better.  

In the low and medium settings, BaggFS largely performs as expected with respect to classical FS.  At low SNRs, the variance stabilization offered by BaggFS allows it to outperform FS but that advantage dies out at medium SNRs and both procedures appear similarly optimal relative to the others at high SNRs.  It can also be seen in these settings that as originally hypothesized above, RandFS with fixed $\mtry = 0.33$ outperforms BaggFS at low SNRs but loses the advantage at high SNRs.  

The performance of RandFS in these settings with respect to the other procedures, however, is what is perhaps most surprising.  Note that in the low setting, the RandFS procedure with tuned $\mtry$ outperforms all other methods -- including lasso and relaxed lasso -- until the mid-range SNR values.  In the medium setting, RandFS exhibits a similar property to that of relaxed lasso, performing very well at low SNRs but adapting (likely selecting larger values of $\mtry$) to also perform very well at large SNRs.  Even more remarkable is the performance of RandFS when $\mtry = 0.33$ and is not tuned -- in both the low and medium settings, even this default RandFS outperforms the lasso across all SNRs.

\subsection{Randomization as Implicit Shrinkage}
Before concluding our work, we provide some additional intuition into the implicit regularization that appears to be taking place with the randomized ensembles (random forests and RandFS) studied in previous sections.  This is more apparent and easily described in the more classical linear model forward selection setting with RandFS, though the same kind of effect is likely present with random forests, which might be seen as simply a more complex form of randomized forward selection.  

As above, suppose we have data of the form $\mathcal{D}_n = \{Z_1, ..., Z_n\}$ where each $Z_i = (\bm{X}_i,Y_i)$, $\bm{X}_i = (X_{1,i}, ..., X_{p,i})$ denotes a vector of $p$ features, $Y \in \mathbb{R}$ denotes the response, and the variables have a general relationship of the form $Y = f(X)+\epsilon$.  Now suppose that we obtain a regression estimate $\hat{f}_{\text{RFS}} = X\hat{\beta}_\text{RFS}$ by averaging over $B$ models, each built via randomized forward selection on $\mathcal{D}_n$ to a depth of $d$.  Note that this is identical to the RandFS procedure described above except that for technical reasons, models are built on the original data each time rather than on bootstrap samples.  Each of the $B$  individual models produces an estimate of the form
\[
\hat{\beta}^{(b)}_{\text{RFS}} = \hat{\beta}_{0}^{(b)} + X_{(1)}^{(b)} \hat{\beta}_{(1)}^{(b)} + \cdots + X_{(d)}^{(b)} \hat{\beta}_{(d)}^{(b)}
\]
\noindent where $X_{(j)}^{(b)}$ is the feature selected at the $j^{th}$ step in the $b^{th}$ model and $\hat{\beta}_{(j)}^{(b)}$ is the corresponding coefficient estimate.  Now consider a particular feature, say $X_1$, and suppose that the ordinary least squares (OLS) estimator exists and that the OLS coefficient estimate for $X_1$ is given by $\hat{\beta}_{1,\text{OLS}}$.  More generally, given an active set $\mathcal{A} \subset \{1, ..., p \}$ containing a subset of feature indices, let $\hat{\beta}_{1,\text{OLS} | \mathcal{A}}$ denote the OLS estimate of the coefficient for $X_1$ when calculated over only the features with indices in $\mathcal{A}$.

Given an orthogonal design matrix, for any indexing set $\mathcal{A}$, $\hat{\beta}_{1,\text{OLS} | \mathcal{A}}$ is equal to $\hat{\beta}_{1,\text{OLS}}$ whenever $1 \in \mathcal{A}$ and equal to 0 otherwise.  Thus, if $\mathcal{A}_b$ denotes the indices of those features selected for inclusion in the $b^{th}$ model, then $\hat{\beta}_{1}^{(b)} = \hat{\beta}_{1,\text{OLS}}$ if $X_1$ is selected (i.e.\ $1 \in \mathcal{A}_b$) and $\hat{\beta}_{1}^{(b)} = 0$ otherwise.  The final coefficient estimate produced by RandFS is thus of the form 
\[
\hat{\beta}_{1,\text{RFS}} = \frac{1}{B} \sum_{i=1}^{B} \beta_{1}^{(b)} = \alpha_1 \cdot \hat{\beta}_{1,\text{OLS}} + (1-\alpha_1) \cdot 0 = \alpha_1 \cdot \hat{\beta}_{1,\text{OLS}}
\]
\noindent where $0 \leq \alpha_1 \leq 1$ denotes the proportion of models in which $X_1$ appears.  In practice, for each feature $X_k$, the selection proportion $\alpha_k$ will depend on the particular data observed, the value of $\mtry$, the depth $d$ of each model, and the importance of $X_k$ relative to the other features.  In particular though, so long as $d$ is relatively large, $\alpha_k$ should still be expected to be somewhat large even for moderately small values of $\mtry$ whenever $X_k$ is a relatively important feature because it will have a very good chance of being included in the model if it is made eligible at any step.  Thus, in this sense, not only does RandFS have a shrinkage effect on each variable, but variables that appear more important by virtue of being included in many models will be shrunken by less than those included only occasionally.  

In the RandFS setup, this varying amount of shrinkage is a by-product of the adaptive, forward-selection nature in which the models are fit.  We now show that when the adaptivity is removed and the linear sub-models are instead fit via standard OLS, the resulting shrinkage becomes more uniform.  In a very recent study released almost simultaneously to this work, \cite{Lejeune2019} demonstrate similar results for these kinds of OLS ensembles; we encourage interested readers to see this work for further results.

Assuming i.i.d.\ data of the form above, suppose that the true relationship between the features and response is given by
\begin{align}
	Y_i = \bm{X}_i'\bm{\beta} +  \epsilon_i  \label{eqt:model}
\end{align}
where $\bm{\epsilon} = (\epsilon_1, ..., \epsilon_n)$ are i.i.d.\ and independent of the original data and each noise term $\epsilon_i$ has mean 0 and variance $\sigma_\epsilon^2$. Denote $\bm{Y} =[Y_i]_{i=1}^n \in \mathbb{R}^n $ and $\bm{X} = [\bm{X}_1, \dots, \bm{X}_n ]' \in \mathbb{R}^{n \times p}$.  Assume $p < n$ so that the OLS estimator on the original data exists. Now suppose that we form a regression estimate $\bt{ens}$ by averaging across $B$ different OLS models, each built using only a subset of $m<p$ features selected uniformly at random.  The following result gives that this average of OLS estimators is equivalent to ridge regression with shrinkage penalty $\lambda = \frac{p - m}{m}$.

\begin{theorem}
Under the data setup given above, assume that $n>p$ and the design matrix $\bm{X}$ is orthogonal.  Then
\begin{align*}
	\bt{ens} \xrightarrow{B \rightarrow \infty} \frac{m}{p} \bt{OLS}
\end{align*}
where $\bt{ens}$ denotes the estimate formed by averaging across $B$ different OLS models, each built using only a subset of $m<p$ features selected uniformly at random, and $\bt{OLS}$ denotes the standard OLS estimate on the original data.
\end{theorem}

\noindent \emph{Proof:}   
For $b = 1 \dots B$, let $S_b \subseteq \{1, \dots, p\} = [p]$ denote the set of indices of the $m$ features selected in the $b^{th}$ model. Let $\bm{S}_b$ be the $p \times m $ subsampling matrix obtained by selecting the columns from $I_p$ corresponding to the indices in $S_b$. The $b^{th}$ model estimate is given by
\[  \bt{(b)} = \bm{S}_b \left(\bm{S}_b'\bm{X}'\bm{X}\bm{S}_b \right)^{-1} \bm{S}_b'\bm{X}'\bm{Y}  \]
which, by orthogonality of $\bm{X}$, can be written as
\[  \bt{(b)} = \bm{S}_b\bm{S}_b'\bm{X}'\bm{Y}= \bm{S}_b\bm{S}_b' (\bm{X}'\bm{X})^{-1} \bm{X}'\bm{Y} = \bm{S}_b\bm{S}_b'\bt{OLS}. \]
Averaging across the $B$ individual models gives
\begin{align}
\bt{ens} = \frac{1}{B} \sum_{b=1}^B \bt{(b)} = \frac{1}{B} \sum_{b=1}^B \bm{S}_b\bm{S}_b' \bt{OLS} . \label{eqt:bt_ens1}  
\end{align}
 Finally, let $\bm{C}$ denote the $p \times p$ diagonal matrix where $\bm{C}_{jj}$ is the number of times that the $j^{th}$ feature is selected in the $B$ base models. Then we have 
\begin{align*}
\bt{ens} = \frac{1}{B}\bm{C}\bt{OLS} \xrightarrow{B \rightarrow \infty} \frac{m}{p} \bt{OLS} .
\end{align*}
  ~ \hfill $\blacksquare$ \\

The above result explicitly shows the shrinkage that occurs when an ensemble estimate is formed by averaging across $B$ models, each of which uses only a randomly selected subset of the available features.  We now demonstrate that when base models are constructed by subsampling both features and observations, a similar result holds in expectation.  \cite{Lejeune2019} showed that the optimal risk of such estimators is equivalent to that of the optimal ridge estimator.  Here we follow the same setup to explicitly examine the expectation of this kind of estimator. 

Assume that the rows of $\bm{X}$ are i.i.d.\ with mean $\bm{0}$ and variance $I_p$  and consider an ensemble of $B$ total OLS base models constructed as follows.  For $b = 1 \dots B$, subsample $m$ features uniformly at random and let $S_b \subseteq \{1, \dots, p\} = [p]$ denote the set of indices corresponding to the features selected in the $b^{th}$ model. Let $\bm{S}_b$ denote the $p \times m $ subsampling matrix obtained by selecting the columns from $I_p$ corresponding to the indices in $S_b$. Similarly, subsample $t$ observations and let $T_b \subseteq [n]$ denote the set of the indices of observations selected in the $b^{th}$ model and let $\bm{T}_b$ denote the $n \times t $ subsampling matrix obtained by selecting the columns from $I_n$ corresponding to the indices in $T_b$. Let $\mathcal{S}$ and $\mathcal{T}$ denote the collections of all possible $S_b$ and $T_b$, respectively.  Finally, assume that all subsampling is carried out independently and define $\bt{ens,ss}$ as the ensemble estimate formed by averaging across the $B$ subsampled OLS models.  The following result gives that on average, $\bt{ens,ss}$ produces an estimate equal to the true coefficient $\bm{\beta}$ shrunk by a factor of $\frac{m}{p}$.

\begin{theorem}
Under the setup given above, assume that $m < t -1$ and $Y_i = \bm{X}_i'\bm{\beta} + $ $\epsilon_i$  as in (\ref{eqt:model}).  Then
\begin{align*}
\mathbb{E} \left( \bt{ens,ss} \right) =  \frac{m}{p}\bm{\beta} = \frac{1}{1 + \frac{p-m}{m} }\bm{\beta} .
\end{align*}
\end{theorem}

\noindent \emph{Proof:}   
For $b = 1 \dots B$, \cite{Lejeune2019} showed that each individual model estimate can be written as 
\[  \bt{(b)} = \bm{S}_b \left(\bm{T}_b'\bm{X}\bm{S}_b \right)^{+} \bm{T}_b'\bm{Y}  \]
where $(\cdot)^{+}$ denotes the Moore-Penrose pseudoinverse. The ensemble estimate is thus given by 
\begin{align}
\bt{ens,ss} = \frac{1}{B} \sum_{b=1}^B \bt{(b)} = \frac{1}{B} \sum_{b=1}^B \bm{S}_b \left(\bm{T}_b'\bm{X}\bm{S}_b \right)^{+} \bm{T}_b'\bm{Y}. \label{eqt:bt_ens2}
\end{align}

Now, given the assumed linear relationship between $Y_i$ and $\bm{X}_i$, the expectation of $\bt{ens,ss}$ with respect to $\bm{\epsilon}$ but conditional on $\bm{X}$, $\mathcal{S}$, and $\mathcal{T}$, is given by
\[ \mathbb{E}_{\bm{\epsilon}} \left( \bt{ens,ss} \right) = \frac{1}{B} \sum_{b=1}^B \bm{S}_b \left(\bm{T}_b'\bm{X}\bm{S}_b \right)^{+} \bm{T}_b'\bm{X}\bm{\beta}.  \]
Applying a further result from \cite{Lejeune2019} showing that $\bm{S}_b\bm{S}_b' + \bm{S}_b^c\bm{S}_b^{c\prime} = I_p$, we have
 \begin{align*}
 \mathbb{E}_{\bm{\epsilon} } \left( \bt{ens,ss} \right) &= \frac{1}{B} \sum_{b=1}^B \bm{S}_b \left(\bm{T}_b'\bm{X}\bm{S}_b \right)^{+} \bm{T}_b'\bm{X}\left( \bm{S}_b\bm{S}_b' + \bm{S}_b^c\bm{S}_b^{c\prime} \right)\bm{\beta} \nonumber	\\
	&= \frac{1}{B} \sum_{b=1}^B  \left( \bm{S}_b\left(\bm{T}_b'\bm{X}\bm{S}_b \right)^{+} \bm{T}_b'\bm{X} \bm{S}_b\bm{S}_b'\bm{\beta} + \bm{S}_b \left(\bm{T}_b'\bm{X}\bm{S}_b \right)^{+} \bm{T}_b'\bm{X} \bm{S}_b^c\bm{S}_b^{c\prime}\bm{\beta} \right) \nonumber	\\
	&= \frac{1}{B} \sum_{b=1}^B  \left( \bm{S}_b\bm{S}_b'\bm{\beta} + \bm{S}_b \left(\bm{T}_b'\bm{X}\bm{S}_b \right)^{+} \bm{T}_b'\bm{X} \bm{S}_b^c\bm{S}_b^{c\prime}\bm{\beta} \right). \label{eqt:bt_ens_epsilon}
\end{align*} 
By the assumption that $\mathbb{E}(\bm{X}_i) = \bm{0}$ and $\Var(\bm{X}_i) = I_p$, $\bm{X}\bm{S}_b$ and $\bm{X}\bm{S}_b^c$ are independent with mean $\bm{0}$. Thus, the expectation of $\bt{ens,ss}$ with respect to the training data $\bm{\epsilon}$ and $\bm{X}$ and conditional on $\mathcal{S}$ and $\mathcal{T}$ is given by 
 \begin{align}
	\mathbb{E}_{\bm{\epsilon}, \bm{X} } \left( \bt{ens,ss} \right) &= \frac{1}{B} \sum_{b=1}^B \bm{S}_b\bm{S}_b'\bm{\beta} = \frac{1}{B} \bm{C}\bm{\beta} 
\end{align}
where $\bm{C}$ is a diagonal matrix with $\bm{C}_{jj}$ equaling the number of times that the $j^{th}$ feature is selected for $j = 1, ..., p$. Since both the features and observations are subsampled uniformly at random, we have
 \begin{align*}
	\mathbb{E}_{\bm{\epsilon}, \bm{X}, \mathcal{S}, \mathcal{T} } \left( \bt{ens,ss} \right) &=  \frac{m}{p}\bm{\beta} = \frac{1}{1 + \frac{p-m}{m} }\bm{\beta} 
\end{align*} 
  as desired. \hfill $\blacksquare$ \\

Note that in both theorems above, the ensemble estimates are shrunk by a factor of $m/p$ and since features are selected uniformly at random (u.a.r.) for each model, this corresponds simply to the probability of each feature being chosen in each model.  As discussed initially, in the RandFS framework where each base model is constructed in an adaptive fashion, the amount of shrinkage applied to each feature will be affected by its relative importance in predicting the response. Indeed, notice from the proofs of both theorems above that if features were selected in any non-u.a.r.\ so that some features were more likely to be selected than others, then so long as those selection probabilities were independent of the original data, only the final lines in the proofs would need changed. In particular, for features more likely to be selected, the expected corresponding diagonal entry of the $\bm{C}$ matrix would be larger, resulting in less shrinkage on the corresponding coefficient.  

\section{Discussion}
\label{sec:discussion}
The results in the previous sections provide substantial evidence that the $\mtry$ parameter -- the distinguishing feature of random forests -- has an implicit regularization effect on the procedure.  Much like the tuning parameter $\lambda$ in explicit regularization procedures like lasso and ridge regression, $\mtry$ serves to mitigate overfitting to particular features.  Thus, contrary to conventional wisdom, random forests are not simply ``better" than bagging, but rather, the relative success of random forests depends heavily on the amount of noise present in the problem at hand.  At low to moderate SNRs, random forests are seen to perform substantially better than bagging whereas bagging becomes far more competitive in high SNR regimes.  This suggests further that at least in regression contexts, the success of random forests is not in fact due to any kind of potential interpolation as was recently hypothesized in \cite{Wyner2017}.  In Section \ref{sec:RFS} we showed that the same kinds of patterns emerged for ensemble-ized extensions of classic forward selection.  Our findings suggest that these modified forward selection procedures intended as bagging and random forest analogues -- BaggFS and RandFS, respectively -- may sometimes provide a substantial improvement over standard forward selection, especially at low SNRs.

The obvious question is then, ``Why do random forests \emph{appear} to work so well in practice on real datasets?"  As discussed in the introduction, there is certainly considerable evidence that random forests do often perform very well in a variety of settings across numerous scientific domains.  In our view, the clear answer is that in practice, many datasets simply \emph{are} quite noisy.  \cite{Hastie2017} provide a thorough discussion along these lines, arguing that although commonly employed in simulations, on real data, SNRs as large as 5 or 6 are extremely rare.  If this is indeed the case, it would explain why random forests are so often viewed as inherently superior to bagging.

Finally, we note that our findings are very much in line both with previous findings and with common practice.  In practical applications, random forests are generally used ``off-the-shelf" without any tuning of the $\mtry$ parameter.  The plots corresponding to the low and medium settings in Figure \ref{fig:randfwd_performance} may shed some light on this:  though not always optimal, the procedure with a fixed $\mtry$ value of 0.33 generally performs quite well and thus, in terms of practical guidance, this default value seems to be as good a starting place as any.  On the other hand, the results provided above do strongly support the notion discussed in many previous studies \citep{Diaz2006,Genuer2008,Bernard2009,Genuer2010,Probst2019} that the $\mtry$ parameter can significantly influence performance.  Just as with explicit regularization procedures, the results above suggest that the $\mtry$ parameter in random forests can be thought of as controlling the amount of shrinkage and regularization and thus is best tuned in practice. For large datasets where tuning could introduce a computational burden, the recent results in \cite{Lopes2019classification} suggest that at least in some cases, the algorithmic variance may quickly diminish after relatively few bootstrap samples and thus the tuning and validation could potentially be done reasonably well even on relatively small ensembles.
 
\newpage
\acks{This research was supported in part by the University of Pittsburgh Center for Research Computing through the resources provided.  LM was partially supported by NSF DMS-2015400.}

\appendix

\section{Additional Simulations and Figures}
\label{app:plots}

In Section \ref{sec:RFdof}, Figure \ref{fig:bagrf_df} shows the estimated degrees of freedom for random forests across various values of $\mtry$ in four different regression setups at a fixed SNR of 3.52.  In each case we see that the dof increases with both $\maxnodes$ and $\mtry$ and we note that the same relationships were seen to also emerge in alternative setups.  Figure \ref{fig:bagrf_df_app} below shows the same experiments carried out with the SNR set to the much smaller level of 0.09 and indeed the findings remain the same. \\

\begin{figure}[!hb]
	\centering
	\includegraphics[width=0.8\textwidth]{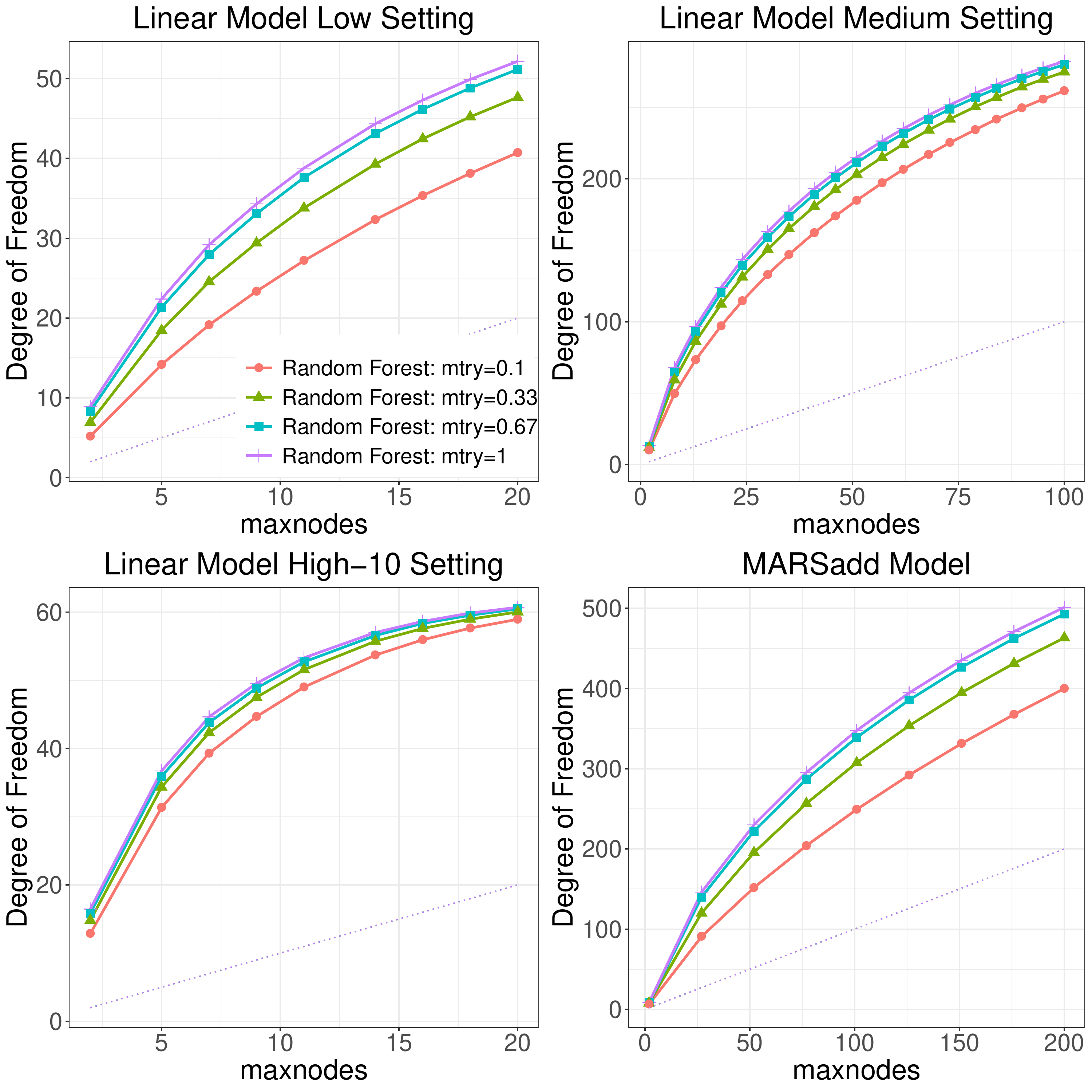}
	\caption{Degrees of freedom for random forests at different levels of \texttt{mtry} with a SNR of 0.09.}
	\label{fig:bagrf_df_app}
\end{figure}

\begin{figure}[!hbp]
	\centering
	\includegraphics[width=0.42\textwidth]{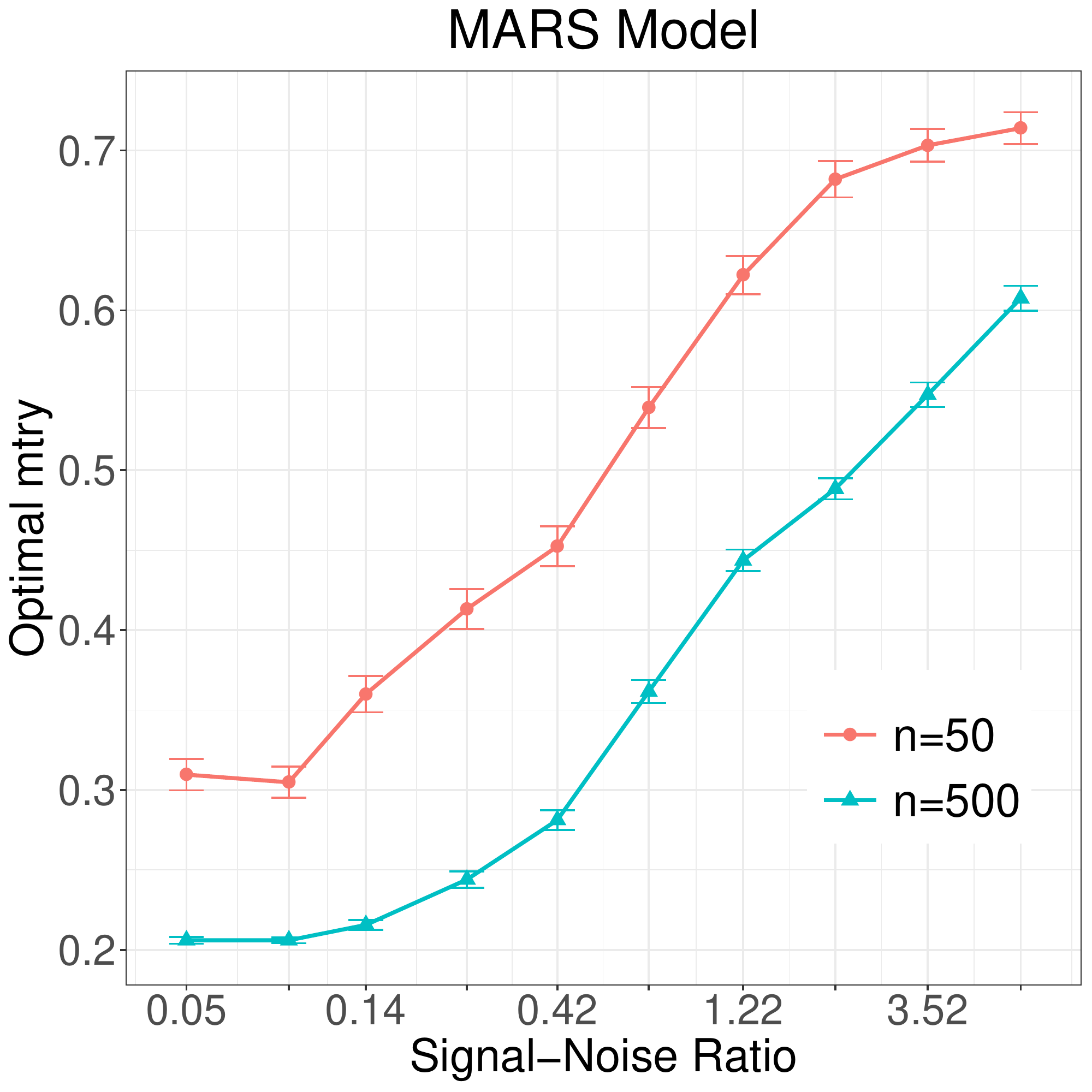}
	\includegraphics[width=0.42\textwidth]{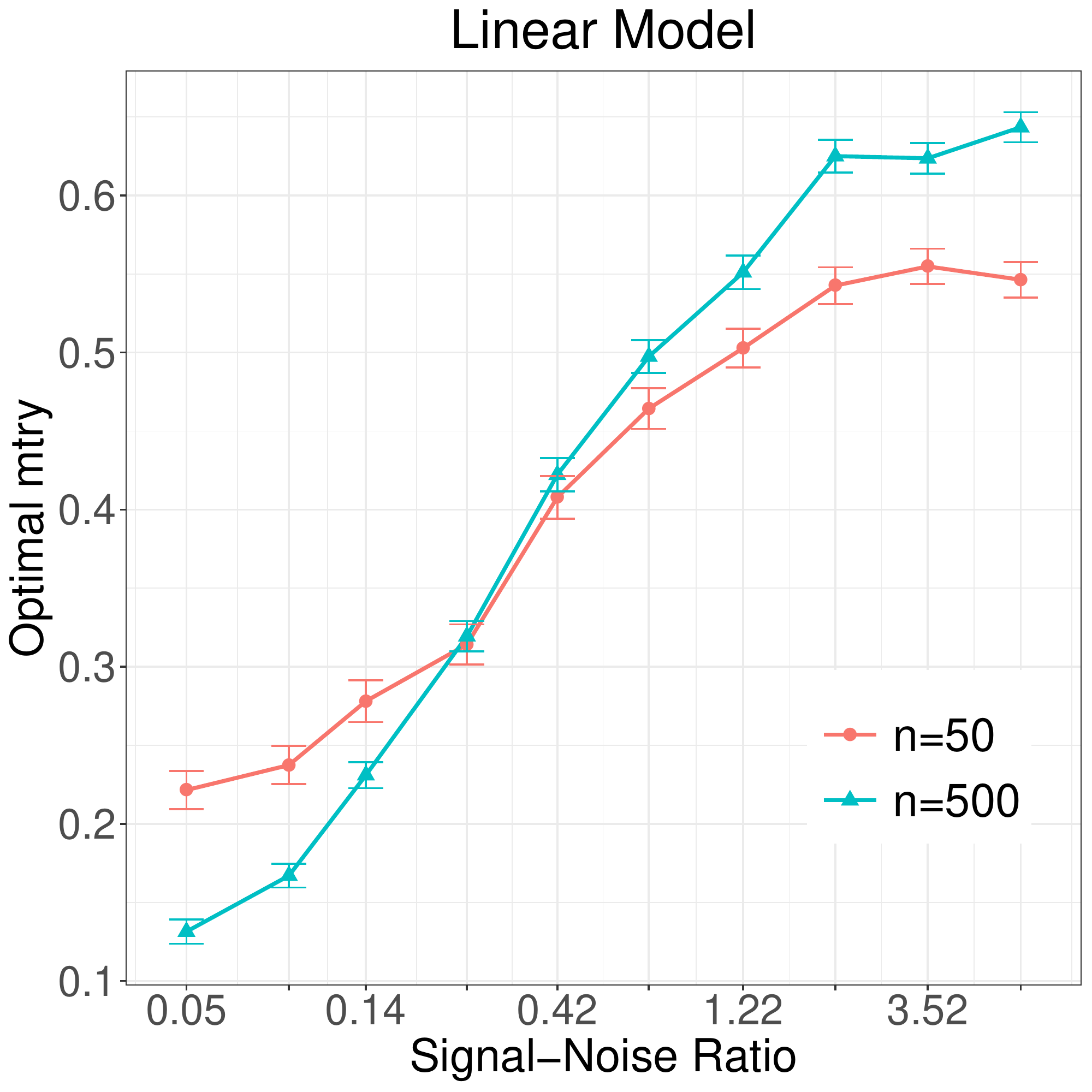}
	\caption{ Optimal value of $\mtry$ vs SNR for the MARS and linear model.}
	\label{fig:optmtrymean}
\end{figure}

\noindent In Section \ref{sec:optmtry} we considered the problem of estimating the optimal value of $\mtry$ for random forests for both the linear and MARS models at various SNR levels.  In the main text, Figure \ref{fig:optmtry} shows plots of the optimal $\mtry$ measured as that which obtains the lowest average test error over 500 iterations.  In contrast, Figure \ref{fig:optmtrymean} here calculates the optimal $\mtry$ value on each of the 500 iterations and then takes the empirical mean.  The results in Figure \ref{fig:optmtrymean} show the same general pattern as seen in Figure \ref{fig:optmtry}.  As the SNR increases, so does the optimal value of $\mtry$. \\

\vspace{10mm}

\vspace{10mm}

\begin{figure}
\centering
\includegraphics[width=0.8\textwidth]{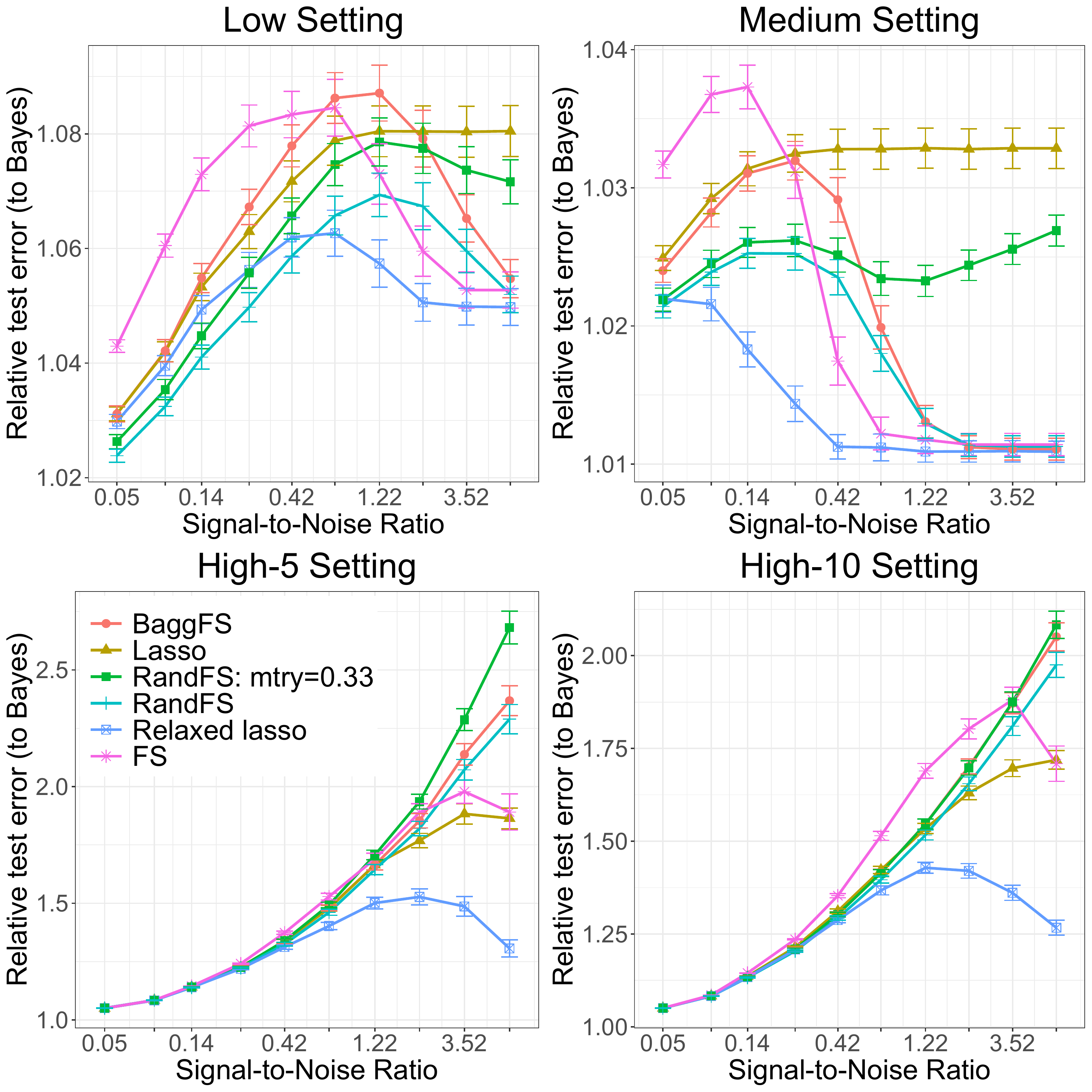}
\caption{Performance of FS, BaggFS, RandFS, lasso and relaxed lasso across SNR levels for linear models in the low, medium, high-5, and high-10 settings. Error bars at each point correspond to $\pm 1$ standard deviation across the 100 replications.}
\label{fig:randfwd_performance_se}
\end{figure}

\noindent Figure \ref{fig:randfwd_performance_se} shows exactly the same plots as in Figure \ref{fig:randfwd_performance} but here we add error bars at each point corresponding to $\pm 1$ standard deviation across the 100 replications.  These are reserved for the appendix only for readability as the error bars can make the plots a bit more muddled and difficult to make out.

\newpage
\FloatBarrier
\bibliography{database}

\end{document}